\documentclass[10pt,twocolumn,letterpaper]{article}

\usepackage[pagenumbers]{wacv} % To force page numbers, e.g. for an arXiv version

\usepackage{graphicx}
\usepackage{amsmath}
\usepackage{amssymb}
\usepackage{booktabs}
\usepackage{amsfonts}
\usepackage{array}
\usepackage{textcomp}
\usepackage{stfloats}
\usepackage{url}
\usepackage{verbatim}
\usepackage{cite}
\usepackage{mathtools}
\usepackage{amsthm}
\usepackage{bbm}
\usepackage{dsfont}
\usepackage{multirow,multicol}
\usepackage[dvipsnames,table]{xcolor}     % colors
\usepackage{siunitx}
\usepackage{footmisc}
\usepackage{booktabs}
\usepackage{microtype}
\usepackage[linesnumbered,ruled,vlined]{algorithm2e}
\usepackage{colortbl}
\usepackage{pifont}
\usepackage{soulutf8}
\colorlet{LightGreen}{SpringGreen!60}
\SetKwInput{KwInput}{Input}                % Set the Input
\SetKwInput{KwOutput}{Output}              % set the Output

\SetCommentSty{mycommfont}

\usepackage[pagebackref,breaklinks,colorlinks]{hyperref}

\usepackage[capitalize]{cleveref}
\crefname{section}{Sec.}{Secs.}
\Crefname{section}{Section}{Sections}
\Crefname{table}{Table}{Tables}
\crefname{table}{Tab.}{Tabs.}

\def\wacvPaperID{1673} % *** Enter the WACV Paper ID here
\def\confName{WACV}
\def\confYear{2024}

\begin{document}

\title{SequenceMatch\\ 
Revisiting the design of weak-strong augmentations for Semi-supervised learning}

\author{Khanh-Binh Nguyen\\
Sungkyunkwan University\\
South Korea\\
{\tt\small n.k.binh00@gmail.com}
% \and
}
\maketitle

%%%%%%%%% ABSTRACT
\begin{abstract}
Semi-supervised learning (SSL) has become popular in recent years because it allows the training of a model using a large amount of unlabeled data.
However, one issue that many SSL methods face is the confirmation bias, which occurs when the model is overfitted to the small labeled training dataset and produces overconfident, incorrect predictions.
To address this issue, we propose SequenceMatch, an efficient SSL method that utilizes multiple data augmentations.
The key element of SequenceMatch is the inclusion of a medium augmentation for unlabeled data.
By taking advantage of different augmentations and the consistency constraints between each pair of augmented examples, SequenceMatch helps reduce the divergence between the prediction distribution of the model for weakly and strongly augmented examples.
In addition, SequenceMatch defines two different consistency constraints for high and low-confidence predictions.
As a result, SequenceMatch is more data-efficient than ReMixMatch, and more time-efficient than both ReMixMatch ($\times4$) and CoMatch ($\times2$) while having higher accuracy.
Despite its simplicity, SequenceMatch consistently outperforms prior methods on standard benchmarks, such as CIFAR-10/100, SVHN, and STL-10.
It also surpasses prior state-of-the-art methods by a large margin on large-scale datasets such as ImageNet, with a 38.46\% error rate. Code is available at https://github.com/beandkay/SequenceMatch.
\end{abstract}

\section{Introduction}
Deep Neural Networks (DNNs) have made significant strides in recent years, achieving an extraordinary performance on many tasks such as image recognition \cite{he2016deep}, speech recognition \cite{amodei2016deep}, and natural language processing \cite{socher2012deep}.
The state-of-the-art performance of DNNs is achieved through supervised learning, which requires labeled data.
The empirical observation shows that training DNNs on larger labeled datasets produces a better performance \cite{hestness2017deep,jozefowicz2016exploring,mahajan2018exploring,radford2019language,raffel2020exploring}.
However, the labeled data is limited in quantity and significantly costly due to the hand-labeling process which must be done by experts.

An impressive approach for training models with a large amount of unlabeled data is semi-supervised learning (SSL).
In recent years, SSL has received much attention due to its advantages in leveraging a large amount of unlabeled data.
Since the unlabeled data can be obtained easily without the need for human labor, using SSL results in comparable performance to the supervised learning methods but with a lower cost.
This success has led to the development of many SSL methods  \cite{berthelot2019mixmatch,Berthelot2020ReMixMatchSL,laine2016temporal,lee2013pseudo,Tarvainen2017MeanTA,Xie2020SelfTrainingWN}. 

\begin{figure}[!ht]
    \centering
    \centerline{
    \includegraphics[width=\linewidth]{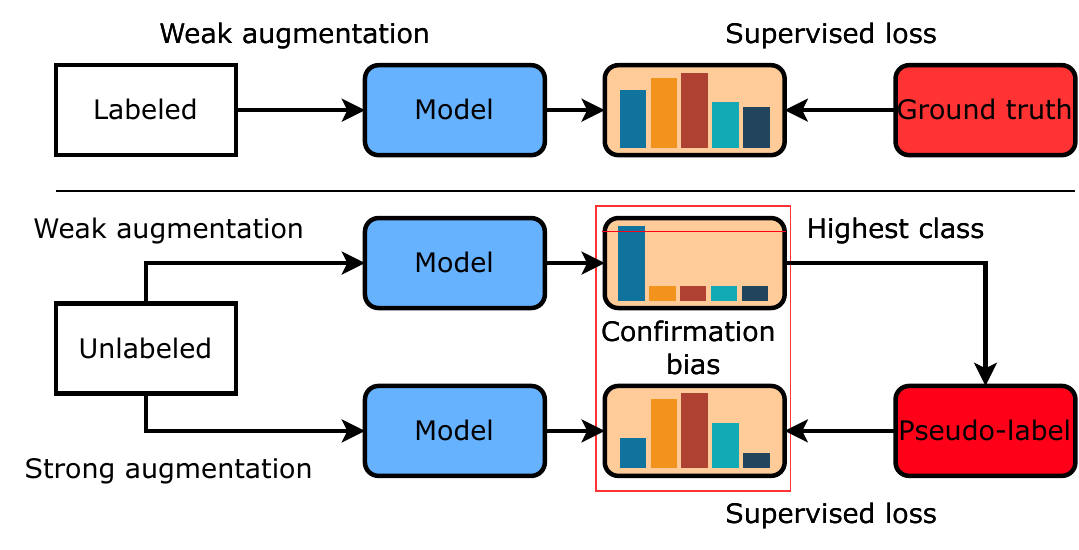}
    }
    \caption{Example scheme where the prediction distribution of weakly and strongly augmented examples have high KL divergence.
    This high divergence happens when the model suffers from the confirmation bias issue.}
    \label{fig:unbalance-distribution}
\end{figure}

\begin{figure*}[!ht]
    \centering
    \resizebox{\linewidth}{!}{
    \includegraphics{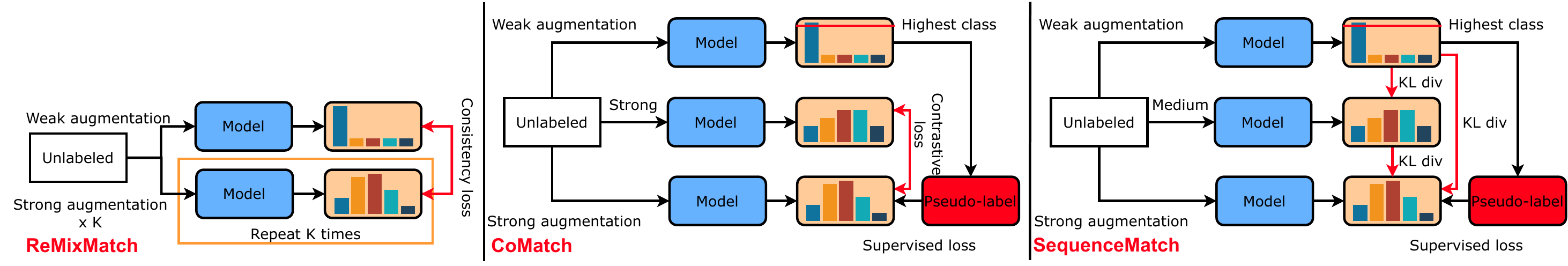}
    }
    \caption{Differences between SequenceMatch versus ReMixMatch and CoMatch methods for multi augmentations.}
    \label{fig:pipeline-diff}
\end{figure*}

There are two popular SSL methods which are pseudo-labeling \cite{lee2013pseudo} (also called self-training \cite{26629,Xie2020SelfTrainingWN}) and consistency regularization \cite{Bachman2014LearningWP,laine2016temporal,Sajjadi2016RegularizationWS}.
While the pseudo-labeling-based approaches use model predictions as labels to train the unlabeled data, the consistency-regularization-based approaches use loss functions such as mean squared error (MSE) or Kullback-Leibler divergence (KL divergence) to minimize the difference between the prediction distribution of different augmented inputs.
However, they are still encountering the confirmation bias issue because of the small labeled training dataset.
Hence, during training, when a confirmation bias issue occurs, the performance stops improving and could become worse.

Based on the finding from \cite{botev2022regularising} that we could utilize KL divergence with multiple augmentations to increase model invariance and generalization, we propose a simple SSL pipeline, SequenceMatch.
The idea of using multiple data augmentations for SSL is not new since it has been introduced by \cite{Berthelot2020ReMixMatchSL} and \cite{li2021comatch}.
ReMixMatch \cite{Berthelot2020ReMixMatchSL} uses a technique called Augmentation Anchoring (AA).
AA anchors a weak augmentation then makes $\mathbf{K}$ strong augmentations and encourages each output to be close to the anchoring prediction.
Similarly, CoMatch \cite{li2021comatch} generates two strongly augmented versions for each unlabeled sample to construct the embedding graph.
However, we argue that using multiple strong augmentations can result in disparate predictions, and thus may not be a meaningful target.
Particularly, ReMixMatch {\cite{Berthelot2020ReMixMatchSL}} found that using stronger augmentations in MixMatch resulted in high divergence, and the training would not converge if we replace the weak with strong augmentation, resulting in very poor performance.
SequenceMatch also uses multiple data augmentations but in a different manner.
Specifically, we introduce a medium augmentation, then minimize the KL divergence between prediction distributions for each pair of inputs, thus minimizing the discrepancy between the representation of weak and strong augmented predictions.
Therefore, by minimizing these divergences, we assume that the learned representation of the strong augmentation would align with the one from the weak augmentation by using medium augmentation as an anchor.
The medium augmentation also works like a Teacher Assistant (TA) to distill the knowledge, similar to a TA in \cite{mirzadeh2020improved}.
As a result, SequenceMatch encourages the similarity of the network outputs to produce more reliable pseudo-labels for unlabeled data during training, reduces overconfident pseudo-labels, and optimizes data utilization for the unlabeled dataset.

The benefit of SequenceMatch is found in all datasets.
For instance, on the STL-10 dataset, SequenceMatch achieves 15.45\%, and 5.56\% error rates when the label amount is 40, and 1000, respectively.
Moreover, SequenceMatch shows its superiority on imbalanced datasets such as SVHN and ImageNet.
On the SVHN dataset, SequenceMatch achieves 1.96\%, 1.89\%, and 1.79\% error rate when the label amount is 40, 250, and 1000, respectively.
For ImageNet, SequenceMatch achieves 38.46\% error rate, surpassing FlexMatch of 41.85\%, FixMatch of 43.66\%, CoMatch of 42.17\%, and FreeMatch of 40.57\%.
In addition, SequenceMatch achieves high performance even though it does not introduce as many augmentations as ReMixMatch and does not need to store the embedded graph like CoMatch.
To sum up, this paper makes the following contributions:
\begin{itemize}
    \item We propose SequenceMatch, a SSL training pipeline that helps reduce the divergence between the prediction distributions of different augmented versions of the same input.
    Therefore, SequenceMatch helps reduce the overconfident predictions and the distribution discrepancy between weakly and strongly augmented predictions.
    \item SequenceMatch leverages the whole unlabeled dataset, including high-confidence and low-confidence predictions, thus optimizing the data utilization.
    \item We verify our hypothesis that reducing the confirmation bias issue of the trained model and reducing the divergence between the prediction distributions would yield better results.
    Hence, SequenceMatch significantly achieves state-of-the-art results on many datasets with different numbers of labels.
\end{itemize}

\section{Analysis of high-confidence and low-confidence pseudo-label}
In order to examine the importance of low-confidence predictions in the training process, we train FixMatch separately with "hard" and "soft" pseudo-labels.
The "hard" pseudo-label training is the conventional FixMatch using high-confidence predictions, while for the "soft" pseudo-label training, the model is trained only on low-confidence predictions.
Specifically, instead of choosing high-confidence predictions as the pseudo-label, we take the low-confidence predictions from weakly-augmented examples, sharpen them by temperature $\mathbf{T=0.5}$ and compute the KL divergence with the predictions from strongly-augmented.

\begin{table}[!h]
    \caption{Error rate of FixMatch using high-confidence vs low-confidence predictions on CIFAR-10 with 40, 250, and 1000-label split.}
    \label{table:high-low}
    \centering
    \resizebox{\linewidth}{!}{%
    \begin{sc}
        \begin{tabular}{lcc} \toprule   \toprule
        Dataset             & high-confidence    & low-confidence     \\ \midrule   \midrule
        CIFAR-10-40         & 7.47          & 28.88         \\
        CIFAR-10-250        & 4.86          & 8.07          \\
        CIFAR-10-4000       & 4.21          & 8.04          \\          \bottomrule \bottomrule
        \end{tabular}
    \end{sc}
    }
\end{table}

The experiment results from Table \ref{table:high-low} show that using only low-confidence predictions to train the model can still achieve a competitive performance with the one using high-confidence predictions on the CIFAR-10 dataset.
This shows that the conventional approach of using a high threshold and discarding a large proportion of unlabeled data during training is inefficient and does not fully leverage the unlabeled data.
Thus, instead of using only high-confidence predictions, in this work, we bridge the strengths of both high-confidence and low-confidence predictions.

\section{Background}
We give a brief introduction to unsupervised data augmentation (UDA) \cite{xie2020unsupervised} and FixMatch \cite{sohn2020fixmatch}, which are mostly related to our work.
Let $B$ be the batch size of labeled data, $\mu$ be the ratio of unlabeled data to labeled data, and $p_{m}$ represent the output probability of the model. 
$A_w$ and $A_{s}$ are weakly and strongly augmentation functions, respectively.
The unsupervised loss term in UDA is formulated as:
\begin{equation}
\begin{aligned}
    \frac{1}{\mu B} \sum_{b=1}^{\mu B} \mathbbm{1}\left(\max \left(q_{b}\right)\geq\tau\right)
    H\left(q_{s}, p_{m}\left(y \mid \mathcal{A}_{s}\left(u_{b}\right)\right)\right),
\end{aligned}
\end{equation}
where $\tau$ is the constant pre-defined threshold, $q_{s} = \frac{\exp{\left(q_{b}/\mathbf{T}\right)}}{\sum_{k}\exp{\left(q_{k}/\mathbf{T}\right)}}$ is the sharpen predictions by temperature $\mathbf{T}$, $q_{b} = p_{m}\left(y \mid \mathcal{A}_{w}\left(u_{b}\right)\right)$ is the logit of label $y$ for input $\mathcal{A}_w\left(u_{b}\right)$.
Unlike UDA, FixMatch leverages this consistency regularization with strong augmentation to achieve a competitive performance.
The unsupervised loss term becomes:
\begin{equation}
\begin{aligned}
    \frac{1}{\mu B} \sum_{b=1}^{\mu B} \mathbbm{1}\left(\max \left(q_{b}\right)\geq\tau\right) H\left(\hat{q}_{b}, p_{m}\left(y \mid \mathcal{A}_{s}\left(u_{b}\right)\right)\right),
\end{aligned}
\end{equation}
where $\hat{q}_{b}=\arg\max\left(q_{b}\right)$ is the pseudo-label of $q_{b}$.
Following FixMatch, FlexMatch uses the same loss term with a dynamic threshold $\tau_{t}$ for each class, thus improving the per-class sampling rate and making the model learn equally.

FixMatch shows that using a high-confidence threshold with "hard" labels can eliminate the noise pseudo-labels, thus enhancing the performance of the whole SSL framework.
In addition, FixMatch also claims that using the high-confidence threshold with the "soft" pseudo-labels does not show a significant difference in performance.

\section{SequenceMatch}
We propose SequenceMatch, a simple SSL pipeline that aims at balancing the prediction distribution of unlabeled data.
The distinction with FixMatch is that we consider both "hard" and "soft" pseudo-labels.
The main novelty comes from the additional medium augmentations for unlabeled data.
With the additional mediumly augmented examples, SequenceMatch helps reduce the divergence between the prediction distributions of the weakly and strongly augmented data.

The intuition is to make the prediction distribution of weakly, mediumly, and strongly augmented examples similar to each other while maintaining the correct pseudo-labels, thus reducing the overfitting of the model on labeled data and reducing the confirmation bias issue.
The medium augmentation is made up of weak augmentation, a transformation chosen at random from the list of strong augmentation transformations, and cutout \cite{devries2017improved}. 
This makes the mediumly augmented samples look different from the weakly augmented ones, but not as distorted as the strongly augmented ones because the induced distortions could severely change the image structures, and thus the transformed images cannot maintain the identity of the original instances.
We visualize the differences of three kinds of augmentation in Appendix \ref{apx:augmentation}.

\subsection{SequenceMatch Pipeline}
\begin{figure}[!ht]
    \centering
    \resizebox{\linewidth}{!}{
    \includegraphics{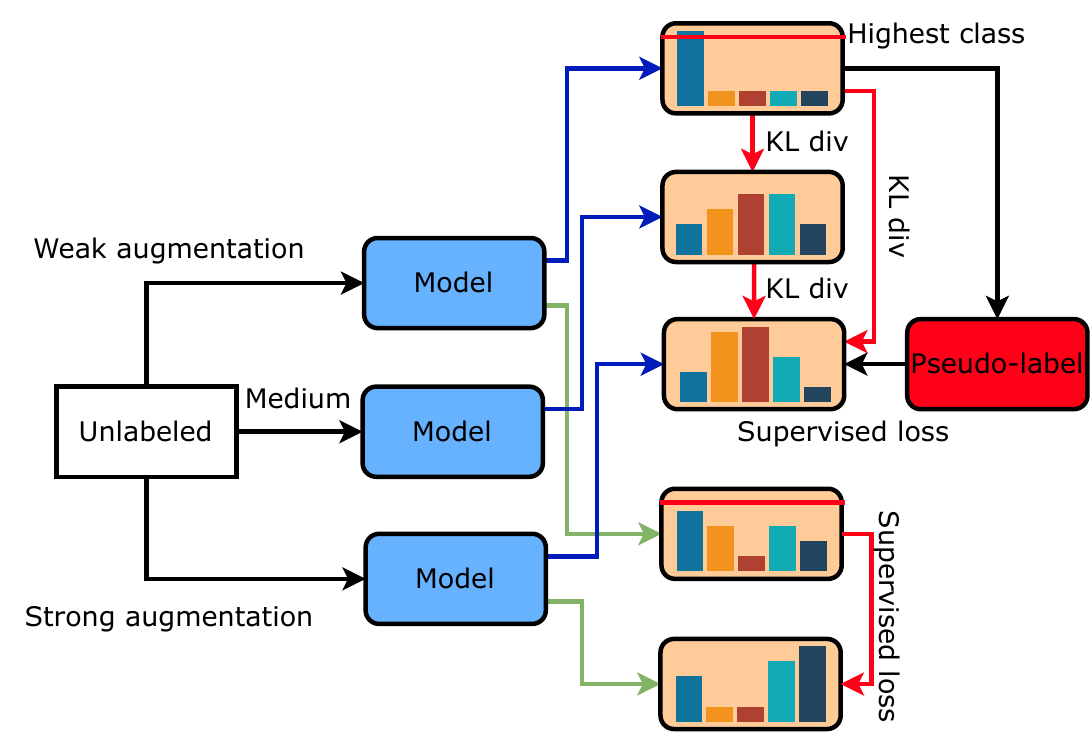}
    }
    \caption{SequenceMatch pipeline. 
    Unlike other SSL methods that use only two types of augmented versions for unlabeled data, we propose a "mediumly augmented" version for unlabeled data.
    The \textcolor{blue}{blue} and \textcolor{green}{green} arrows indicate high-confidence and low-confidence predictions, respectively.
    In addition, we measure the Kullback-Leibler divergence losses between the weakly, mediumly, and strongly augmented versions of the same input, then we minimize them during training.}
    \label{fig:pipeline}
\end{figure}

In this section, we present the pipeline of SequenceMatch method as shown in Figure \ref{fig:pipeline}.
First, similar to other SSL methods, we train the model on the labeled data.
Then, for the unlabeled data, instead of using only weakly and strongly augmented examples, we create three versions of augmented input: weakly, mediumly, and strongly augmented examples.
Finally, for each pair of the prediction distribution such as weak-medium, medium-strong, and weak-strong, a Kullback-Leibler divergence loss function is used to measure the divergence of each pair.
The KL divergence losses will be optimized during the training process to minimize the divergence.

\subsection{Loss Function}
The loss function for SequenceMatch consists of two different loss terms.
One is the supervised loss, which is a standard cross-entropy loss ($\mathcal{L}_{s}^\mathrm{CE}$) for the labeled data.
The other one is the unsupervised loss, including the Kullback-Leibler divergence ($\mathcal{L}_{KL}$) between the prediction distributions and the standard cross-entropy loss ($\mathcal{L}_{u}^\mathrm{CE}$) for strongly augmented data with pseudo-labels.

For an $L$-class classification problem, let $\mathcal{X}=\left\{\left(x_{b}, y_{b}\right): b \in(1, \ldots, B)\right\}$ be a batch of $B$ labeled examples, where $x_{b}$ is the training examples and $y_{b}$ is one-hot labels. 
Let $\mathcal{U}=\left\{u_{b}: b \in \left(1,\dots,\mu{B}\right)\right\}$ be a batch of $\mu{B}$ unlabeled examples where $\mu$ is a hyperparameter that determines the relative sizes of $\mathcal{X}$ and  $\mathcal{U}$. 
Let $p_{m}(y|x)$ is the predicted class distribution of the model for input $x$, $H(p,q)$ denotes the "hard" label cross entropy between two probability distributions $p$ and $q$.
The loss function for SSL is defined as:
\begin{equation}
    \mathcal{L}_\mathrm{SSL}=\mathcal{L}_{s}^\mathrm{CE} + \lambda_u\mathcal{L}_{u},
\end{equation}
where $\lambda_u$ is the fixed weight for the unlabeled data loss.
Specifically, $\mathcal{L}_{s}^\mathrm{CE}$ is a standard cross entropy loss on weakly augmented labeled data:
\begin{equation}
    \mathcal{L}_{s}^\mathrm{CE}=\frac{1}{B} \sum_{b=1}^{B} \mathrm{H}\left(y_{b}, p_{m}\left(y \mid \mathcal{A}_w\left(x_{b}\right)\right)\right)
\end{equation}
Then, let $\mathcal{A}_w$, $\mathcal{A}_{m}$, $\mathcal{A}_{s}$ be the weakly, mediumly, and strongly augmentations for unlabeled data.
$\mathcal{L}_{u}$ is defined as a total of the standard cross-entropy loss ($\mathcal{L}_{u}^\mathrm{CE}$) and the Kullback-Leibler divergence loss ($\mathcal{L}_\mathrm{KL}$).
$\mathcal{L}_{u}^\mathrm{CE}$ has two parts: first is the cross-entropy loss between pseudo-labels and the strongly augmented predictions; second is the cross-entropy loss between sharpen predictions of weakly and strongly augmented samples.
$\mathcal{L}_{KL}$ is the KL divergence of the prediction distribution between each pair of augmented examples $\mathcal{A}_w - \mathcal{A}_{m}$, $\mathcal{A}_w - \mathcal{A}_{s}$, $\mathcal{A}_{m} - \mathcal{A}_{s}$:
\begin{equation}
\begin{aligned}
    &\mathcal{L}_{u} =\mathcal{L}_{u}^\mathrm{CE}+\mathcal{L}_\mathrm{KL}^{w-m}+\mathcal{L}_\mathrm{KL}^{m-s}+\mathcal{L}_\mathrm{KL}^{w-s}
\end{aligned}
\end{equation}
\begin{equation}
\begin{aligned}
    &\mathcal{L}_{u}^\mathrm{CE} = \frac{1}{\mu B} \sum_{b=1}^{\mu B}\left(\mathbbm{1}\left(\max \left(q_{b}^w\right) \geq \tau\right) \mathrm{H}\left(\hat{q}_{b}, p_{m}\left(y \mid \mathcal{A}_{s}\left(u_{b}\right)\right)\right)\right.\\
    &+\left.\mathbbm{1}\left(\max \left(q_{b}^w\right) < \tau\right)\mathrm{H}\left(q_s\mid p_{m}\left(y\mid \mathcal{A}_{s}\left(u_{b}\right)\right)\right)\right)
\end{aligned}
\end{equation}
\begin{equation}
    \footnotesize{\mathcal{L}_\mathrm{KL}^{w-m} = \frac{1}{\mu B} \sum_{b=1}^{\mu B}\mathbbm{1}\left(\max \left(q_{b}^{w}\right) \geq \tau\right) D_\mathrm{KL}\left(q_{s}^{w} \mid p_{m}\left(y \mid \mathcal{A}_{m}\left(u_{b}\right)\right)\right)}
\end{equation}
\begin{equation}
    \footnotesize{\mathcal{L}_\mathrm{KL}^{m-s} = \frac{1}{\mu B} \sum_{b=1}^{\mu B}\mathbbm{1}\left(\max \left(q_{b}^{m}\right) \geq \tau\right) D_\mathrm{KL}\left(q_{s}^{m} \mid p_{m}\left(y \mid \mathcal{A}_{s}\left(u_{b}\right)\right)\right)}
\end{equation}
\begin{equation}
    \footnotesize{\mathcal{L}_\mathrm{KL}^{w-s} = \frac{1}{\mu B} \sum_{b=1}^{\mu B}\mathbbm{1}\left(\max \left(q_{b}^{w}\right) \geq \tau\right) D_\mathrm{KL}\left(q_{s}^{w} \mid p_{m}\left(y \mid \mathcal{A}_{s}\left(u_{b}\right)\right)\right)}
\end{equation}
where $\hat{q}_{b}=\arg\max\left(q_{b}\right)$ is the pseudo-label with $q_{b}=p_{m}\left(y \mid \Omega\left(u_{b}\right)\right)$, $\Omega$ is the corresponding augmentation function, $q_{s} = \frac{\exp{\left(q_{b}/\mathbf{T}\right)}}{\sum_{k}\exp{\left(q_{k}/\mathbf{T}\right)}}$ is the sharpen predictions, $\tau$ is the fixed threshold for choosing pseudo-labels, $D_\mathrm{KL}$ denotes the KL divergence function, and T is the temperature for sharpening.
We use the fixed $\tau$ for high-confidence KL loss to reduce the divergence of overconfident predictions over three augmentations.
Notably, we only enforce the consistency loss for low-confidence on weak-strong predictions pair since the low-confidence predictions from medium and strong augmented predictions are unreliable.
Compared with prior methods, the use of unlabeled data by KL loss is more reasonable, as KL loss will not bring negative supervisory information due to the wrong predictions but just emphasize the distribution consistency between the weakly, mediumly, and strongly augmented images.

\section{Experiments}
We evaluate SequenceMatch on common datasets: CIFAR-10/100 \cite{krizhevsky2009learning}, SVHN \cite{netzer2011reading}, STL-10 \cite{coates2011analysis}, and ImageNet \cite{Deng2009ImageNetAL}, and extensively investigate the performance under various labeled data amounts. 
We mainly compare our proposed method with fully SSL methods without using self-supervised pre-trained weights such as UDA \cite{xie2020unsupervised}, FixMatch \cite{sohn2020fixmatch}, FlexMatch \cite{zhang2021flexmatch}, FreeMatch \cite{wang2022freematch}, etc, since they all include a pre-defined threshold, and they are currently the state-of-the-art in the field. 
To further understand the findings of SSL techniques, we additionally include a fully-supervised experiment for each dataset.
We implement our proposed method and evaluate all methods using USB framework\footnote{https://github.com/microsoft/Semi-supervised-learning/}.

For a fair comparison, we use the same hyperparameters of the UDA, FixMatch, and FlexMatch methods. 
Standard stochastic gradient descent (SGD) with a momentum of 0.9 is used as an optimizer in all experiments \cite{sutskever2013importance,polyak1964some}. 
For all datasets, we use an initial learning rate of 0.03 with a cosine annealing learning rate scheduler \cite{loshchilov2016sgdr} for a total of $2^{20}$ training iterations. 
We also conduct an exponential moving average with a momentum of 0.999. 
The batch size of labeled data is 64, except for ImageNet. 
For CIFAR-10/100, SVHN, and STL-10, $\mu$ is set to 7, and it is set to 1 for ImageNet.
For UDA, $\tau$ is set to 0.8, while it is set to 0.95 for FixMatch, FlexMatch, and SequenceMatch.
These configurations follow the original papers \cite{xie2020unsupervised,sohn2020fixmatch,zhang2021flexmatch}.
The medium and strong augmentation in our experiments is RandAugment \cite{Cubuk2020RandaugmentPA} with a different number of augmentations (1 for medium augmentation and 3 for strong augmentation; we study the choices for medium augmentation and visualize the differences in the Appendix). 
We use ResNet-50 \cite{krizhevsky2009learning} for the ImageNet dataset and Wide-ResNet (WRN) \cite{Zagoruyko2016WideRN} for other datasets.

\subsection{CIFAR-10/100, STL-10, SVHN}
We evaluate the best error rate of each method by averaging results from five runs with distinct random seeds.
The classification error rates on CIFAR-10/100, STL-10, and SVHN datasets are recorded in Table \ref{table:results1}.

\begin{table*}[!ht]
\centering
\caption{Error rates on CIFAR-10/100, SVHN, and STL-10 datasets on 5 different folds.}
\resizebox{\textwidth}{!}{%
\begin{sc}
\begin{tabular}{l|ccc|ccc|ccc|cc}
\toprule \toprule
Dataset & \multicolumn{3}{c|}{CIFAR-10} & \multicolumn{3}{c|}{CIFAR-100} & \multicolumn{3}{c|}{SVHN} & \multicolumn{2}{c}{STL-10} \\ \midrule
\# Label    &   40  &   250 &   4000    &   400 &   2500    &   10000   &   40  &   250 &   1000    &   40  &   1000\\
\midrule \midrule
$\Pi$ Model \cite{rasmus2015semi}   & 74.34\scriptsize{$\pm$1.76}    & 46.24\scriptsize{$\pm$1.29}    & 13.13\scriptsize{$\pm$0.59}    & 86.96\scriptsize{$\pm$0.80}    & 58.80\scriptsize{$\pm$0.66}    & 36.65\scriptsize{$\pm$0.00}    & 67.48\scriptsize{$\pm$0.95}    & 13.30\scriptsize{$\pm$1.12}    & 7.16\scriptsize{$\pm$0.11} & 74.31\scriptsize{$\pm$0.85}    & 32.78\scriptsize{$\pm$0.40}\\
Pseudo Label \cite{lee2013pseudo}   & 74.61\scriptsize{$\pm$0.26}    & 46.49\scriptsize{$\pm$2.20}    & 15.08\scriptsize{$\pm$0.19}    & 87.45\scriptsize{$\pm$0.85}    & 57.74\scriptsize{$\pm$0.28}    & 36.55\scriptsize{$\pm$0.24}    & 64.61\scriptsize{$\pm$5.60} & 15.59\scriptsize{$\pm$0.95}    & 9.40\scriptsize{$\pm$0.32} & 74.68\scriptsize{$\pm$0.99}    & 32.64\scriptsize{$\pm$0.71}\\
VAT \cite{Miyato2019VirtualAT}  & 74.66\scriptsize{$\pm$2.12}    & 41.03\scriptsize{$\pm$1.79}    & 10.51\scriptsize{$\pm$0.12}    & 85.20\scriptsize{$\pm$1.40}    & 46.84\scriptsize{$\pm$0.79}    & 32.14\scriptsize{$\pm$0.19}    & 74.75\scriptsize{$\pm$3.38}    & 4.33\scriptsize{$\pm$0.12} & 4.11\scriptsize{$\pm$0.20} & 74.74\scriptsize{$\pm$0.38}    & 37.95\scriptsize{$\pm$1.12}\\
MeanTeacher \cite{Tarvainen2017MeanTA}  & 70.09\scriptsize{$\pm$1.60}    & 37.46\scriptsize{$\pm$3.30}    & 8.10\scriptsize{$\pm$0.21} & 81.11\scriptsize{$\pm$1.44}    & 45.17\scriptsize{$\pm$1.06}    & 31.75\scriptsize{$\pm$0.23}    & 36.09\scriptsize{$\pm$3.98}    & 3.45\scriptsize{$\pm$0.03} & 3.27\scriptsize{$\pm$0.05} & 71.72\scriptsize{$\pm$1.45}    & 33.90\scriptsize{$\pm$1.37}\\
MixMatch \cite{berthelot2019mixmatch}   & 36.19\scriptsize{$\pm$6.48}    & 13.63\scriptsize{$\pm$0.59}    & 6.66\scriptsize{$\pm$0.26} & 67.59\scriptsize{$\pm$0.66}    & 39.76\scriptsize{$\pm$0.48}    & 27.78\scriptsize{$\pm$0.29}    & 30.60\scriptsize{$\pm$8.39}    & 4.56\scriptsize{$\pm$0.32} & 3.69\scriptsize{$\pm$0.37} & 54.93\scriptsize{$\pm$0.96}    & 21.70\scriptsize{$\pm$0.68}\\
ReMixMatch \cite{Berthelot2020ReMixMatchSL} & 9.88\scriptsize{$\pm$1.03} & 6.30\scriptsize{$\pm$0.05} & 4.84\scriptsize{$\pm$0.01} & 42.75\scriptsize{$\pm$1.05}    & \underline{26.03\scriptsize{$\pm$0.35}}    & \textbf{20.02\scriptsize{$\pm$0.27}}    & 24.04\scriptsize{$\pm$9.13}    & 6.36\scriptsize{$\pm$0.22} & 5.16\scriptsize{$\pm$0.31} & 32.12\scriptsize{$\pm$6.24}    & 6.74\scriptsize{$\pm$0.14}\\
UDA \cite{xie2020unsupervised}  & 10.62\scriptsize{$\pm$3.75}    & 5.16\scriptsize{$\pm$0.06} & 4.29\scriptsize{$\pm$0.07} & 46.39\scriptsize{$\pm$1.59}    & 27.73\scriptsize{$\pm$0.21}    & 22.49\scriptsize{$\pm$0.23}    & 5.12\scriptsize{$\pm$4.27} & \underline{1.92\scriptsize{$\pm$0.05}} & \underline{1.89\scriptsize{$\pm$0.01}} & 37.42\scriptsize{$\pm$8.44}    & 6.64\scriptsize{$\pm$0.17}\\
FixMatch \cite{sohn2020fixmatch}    & 7.47\scriptsize{$\pm$0.28} & \underline{4.86\scriptsize{$\pm$0.05}} & 4.21\scriptsize{$\pm$0.08} & 46.42\scriptsize{$\pm$0.82}    & 28.03\scriptsize{$\pm$0.16}    & 22.20\scriptsize{$\pm$0.12}    & 3.81\scriptsize{$\pm$1.18} & 2.02\scriptsize{$\pm$0.02} & 1.96\scriptsize{$\pm$0.03} & 35.97\scriptsize{$\pm$4.14}    & 6.25\scriptsize{$\pm$0.33}\\ 
Dash \cite{xu2021dash}  & 8.93\scriptsize{$\pm$3.11} & 5.16\scriptsize{$\pm$0.23} & 4.36\scriptsize{$\pm$0.11} & 44.82\scriptsize{$\pm$0.96}    & 27.15\scriptsize{$\pm$0.22}    & 21.88\scriptsize{$\pm$0.07}    & 2.19\scriptsize{$\pm$0.18} & 2.04\scriptsize{$\pm$0.02} & 1.97\scriptsize{$\pm$0.01} & 34.52\scriptsize{$\pm$4.30}    & 6.39\scriptsize{$\pm$0.56}\\
MPL \cite{pham2021meta} & 6.62\scriptsize{$\pm$0.91} & 5.76\scriptsize{$\pm$0.24} & 4.55\scriptsize{$\pm$0.04} & 46.26\scriptsize{$\pm$1.84}    & 27.71\scriptsize{$\pm$0.19}    & 21.74\scriptsize{$\pm$0.09}    & 9.33\scriptsize{$\pm$8.02} & 2.29\scriptsize{$\pm$0.04} & 2.28\scriptsize{$\pm$0.02} & 35.76\scriptsize{$\pm$4.83}    & 6.66\scriptsize{$\pm$0.00}\\
FlexMatch \cite{zhang2021flexmatch} & 4.97\scriptsize{$\pm$0.06} & 4.98\scriptsize{$\pm$0.09} & 4.19\scriptsize{$\pm$0.01} & 39.94\scriptsize{$\pm$1.62}    & 26.49\scriptsize{$\pm$0.20}    & 21.90\scriptsize{$\pm$0.15}    & 8.19\scriptsize{$\pm$3.20} & 6.59\scriptsize{$\pm$2.29} & 6.72\scriptsize{$\pm$0.30} & 29.15\scriptsize{$\pm$4.16}    & 5.77\scriptsize{$\pm$0.18}\\
FreeMatch \cite{wang2022freematch}  & \underline{4.90\scriptsize{$\pm$0.04}} & 4.88\scriptsize{$\pm$0.18} & \textbf{4.10\scriptsize{$\pm$0.02}} & \underline{37.98\scriptsize{$\pm$0.42}}    & 26.47\scriptsize{$\pm$0.20}    & 21.68\scriptsize{$\pm$0.03}    & \underline{1.97\scriptsize{$\pm$0.02}} & 1.97\scriptsize{$\pm$0.01} & 1.96\scriptsize{$\pm$0.03} & \underline{15.56\scriptsize{$\pm$0.55}}    & \underline{5.63\scriptsize{$\pm$0.15}}\\
\rowcolor{LightGreen}\textbf{SequenceMatch} & \textbf{4.80\scriptsize{$\pm$0.01}} & \textbf{4.75\scriptsize{$\pm$0.05}} & \underline{4.15\scriptsize{$\pm$0.01}} & \textbf{37.86\scriptsize{$\pm$1.07}} & \textbf{25.99\scriptsize{$\pm$0.22}} & \underline{20.10\scriptsize{$\pm$0.04}} & \textbf{1.96\scriptsize{$\pm$0.23}} & \textbf{1.89\scriptsize{$\pm$0.31}} & \textbf{1.79\scriptsize{$\pm$0.02}}  & \textbf{15.45\scriptsize{$\pm$1.40}} & \textbf{5.56\scriptsize{$\pm$0.35}}\\ \midrule
Fully-Supervised & \multicolumn{3}{c|}{4.62\scriptsize{$\pm$0.05}} & \multicolumn{3}{c|}{19.30\scriptsize{$\pm$0.09}} & \multicolumn{3}{c|}{2.13\scriptsize{$\pm$0.02}}  & \multicolumn{2}{c}{None}\\ \bottomrule \bottomrule
\end{tabular}%
\end{sc}
}
\label{table:results1}
\end{table*}

For the CIFAR-10 and SVHN datasets, we use Wide-ResNet-28-2 \cite{Zagoruyko2016WideRN} as a backbone model, Wide-ResNet-28-8 for the CIFAR-100 dataset, and Wide-ResNet-37-2 for the STL-10 dataset.
As shown in Table \ref{table:results1}, the proposed method outperforms all other methods on most datasets with varying numbers of labels.
According to FlexMatch study \cite{zhang2021flexmatch}, FlexMatch performs less favorably on imbalanced datasets such as SVHN.
SequenceMatch, on the other hand, not only achieves high performance across all datasets but also performs well on the SVHN dataset.
This shows that our proposed method has the effect of reducing overfitting, which usually appears when training on a small and imbalanced dataset.

\subsubsection*{Precision, Recall, F1 Score and AUC results on CIFAR-10}
To comprehensively evaluate the performance of all methods in a classification setting, we further report the precision, recall, F1-score, and AUC (area under curve) results on the CIFAR-10 dataset. 
As shown in Table \ref{table:details}, we see that in addition to the reduced error rates, SequenceMatch also has the best performance in precision, recall, F1 score, and AUC.
These metrics, together with error rates (or accuracy), show the strong performance of our proposed method.
% It should be noted that even though Comatch, SimMatch, and AdaMatch have better performance in some cases, they are using the Distribution Alignment (DA) technique.

\begin{table*}[t!]
\centering
% \resizebox{\textwidth}{!}{
    \caption{Precision, recall, F1-score and AUC results on CIFAR-10.}
    \label{table:details}
\begin{sc}
    \begin{tabular}{@{}l|cccc|cccc}
    \toprule \toprule
    Label Amount    & \multicolumn{4}{c}{40 labels}      & \multicolumn{4}{|c}{4000 labels}  \\ \midrule
    Criteria    & Precision & Recall    & F1 Score  & AUC       & Precision & Recall    & F1 Score   & AUC \\ \midrule \midrule
    FixMatch    & 0.9333    & 0.9290    & 0.9278    & 0.9910    & 0.9571    & 0.9571    & 0.9569  & 0.9984      \\ \midrule
    FlexMatch   & 0.9506    & 0.9507    & 0.9506    & 0.9975    & 0.9580    & 0.9581    & 0.9580  & 0.9984       \\ \midrule
    FreeMatch   & 0.9510    & 0.9512    & 0.9510    & -         & 0.9568    & 0.9568    & 0.9567    & -    \\  \midrule
    \rowcolor{LightGreen}\textbf{SequenceMatch}   & \textbf{0.9519}   & \textbf{0.9521}    & \textbf{0.9519}    & \textbf{0.9976}    & \textbf{0.9590}    & \textbf{0.9591}    & \textbf{0.9590}  & \textbf{0.9986}        \\ \bottomrule \bottomrule
    \end{tabular}%
% }
\end{sc}
\end{table*}

\subsubsection*{STL-10 Confusion Matrix}
\begin{figure}[!ht]
    \centering
    \includegraphics[width=\linewidth]{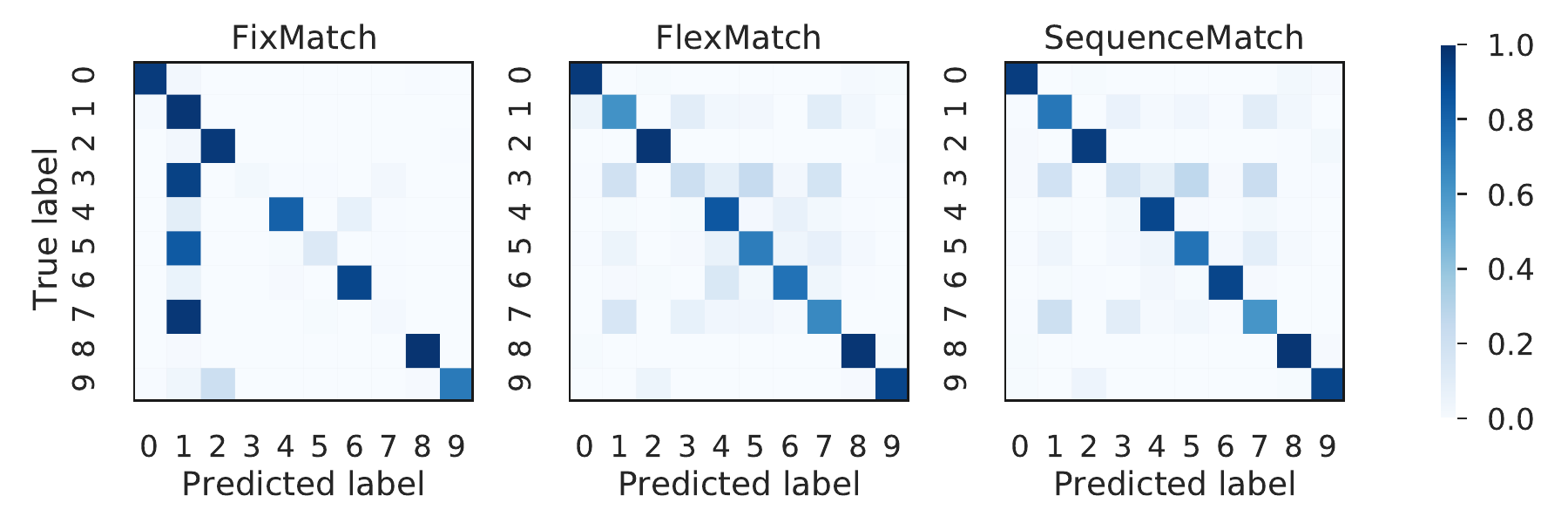}
    \caption{Confusion matrix of FixMatch, FlexMatch, and SequenceMatch features on STL-10 dataset with 40-label.}
    \label{fig:matrix}
\end{figure}

The confusion matrix of FixMatch, FlexMatch, and SequenceMatch on the STL-10 dataset with a 40-label split is visualized in Figure \ref{fig:matrix}.
Compared with FlexMatch, SequenceMatch improves the performance for classes 2, 4, and 6.
In addition, FixMatch is overfitted for class number 1, and it failed to recognize class number 3 and 7.

\subsubsection*{Convergence Speed}
\begin{figure}[!ht]
    \centering
    \subfloat[Top-1 accuracy]
    {\includegraphics[width=.49\linewidth]{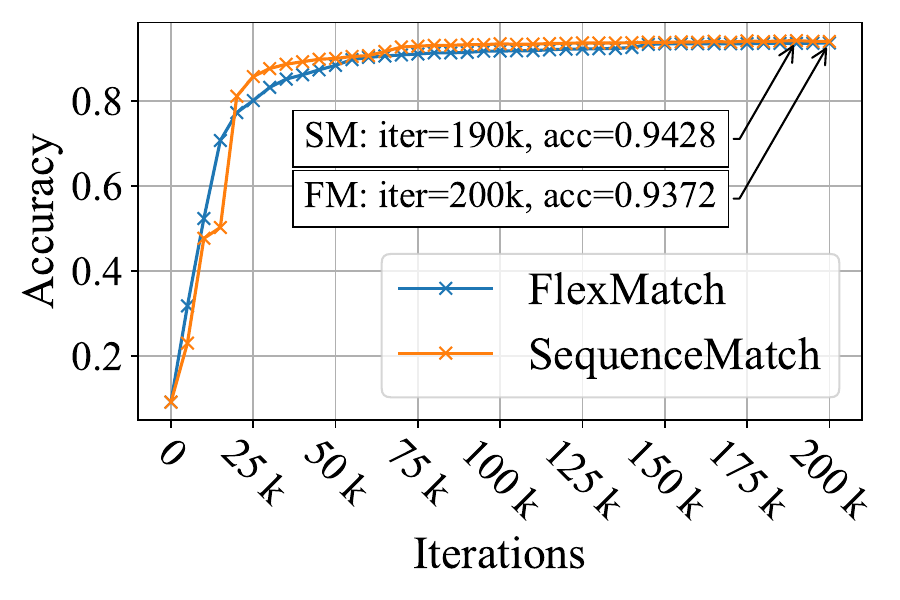}}%
    \subfloat[Loss]
    {\includegraphics[width=.49\linewidth]{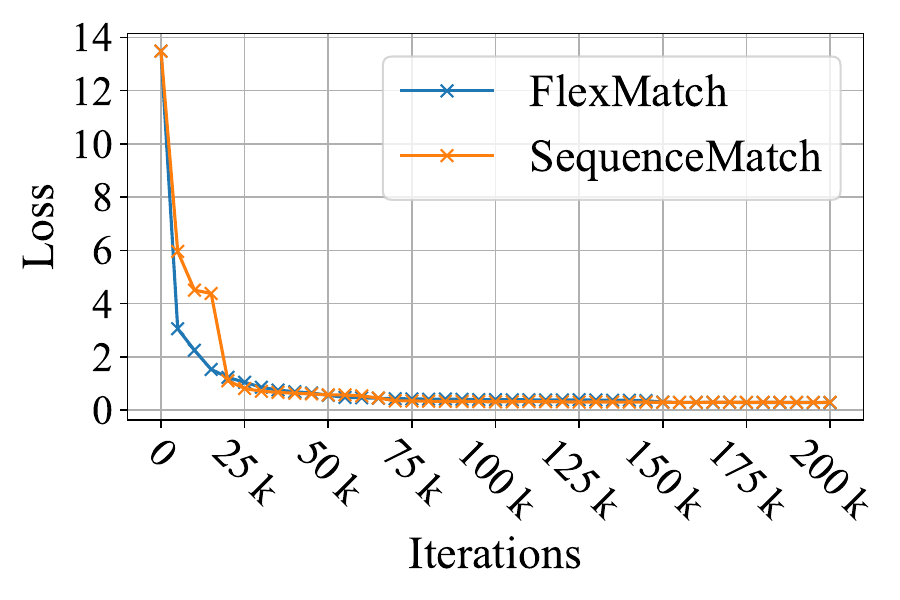}}
    \caption{Convergence analysis vs FlexMatch. (a) and (b) depict the loss and top-1 accuracy on CIFAR-10 with 40 labels. "SM" and "FM" denote SequenceMatch and FlexMatch, respectively.}
    \label{fig:loss_acc}
\end{figure}

Our proposed SequenceMatch outperforms FlexMatch when the number of labels is limited.
We visualize a validation loss and a top-1 accuracy of both FlexMatch and SequenceMatch on the CIFAR-10 dataset with 40 labels within the first 200k iterations.
As we can see in Figure \ref{fig:loss_acc}, SequenceMatch achieves over 80\% of accuracy within the first 25k iterations when FlexMatch nearly hits 80\%. 
After 200k iterations, SequenceMatch achieves up to 94.28\% accuracy while FlexMatch can only achieve 93.72\% of accuracy.
Moreover, the loss of our proposed SequenceMatch also decreases as fast and smoothly as FlexMatch, even though we add extra augmented data.

\subsubsection*{Class-wise accuracy on CIFAR-10 40-label}
\begin{figure}[!ht]
    \centering
    \subfloat{\includegraphics[width=\linewidth]{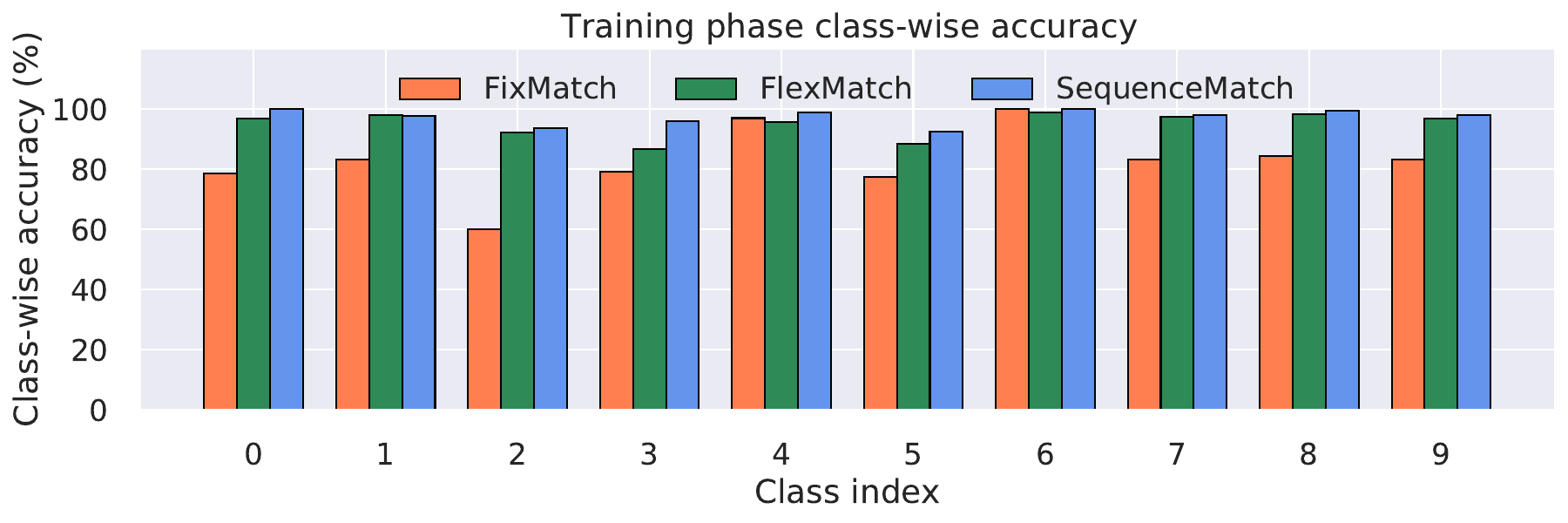}}
    \qquad
    \subfloat{\includegraphics[width=\linewidth]{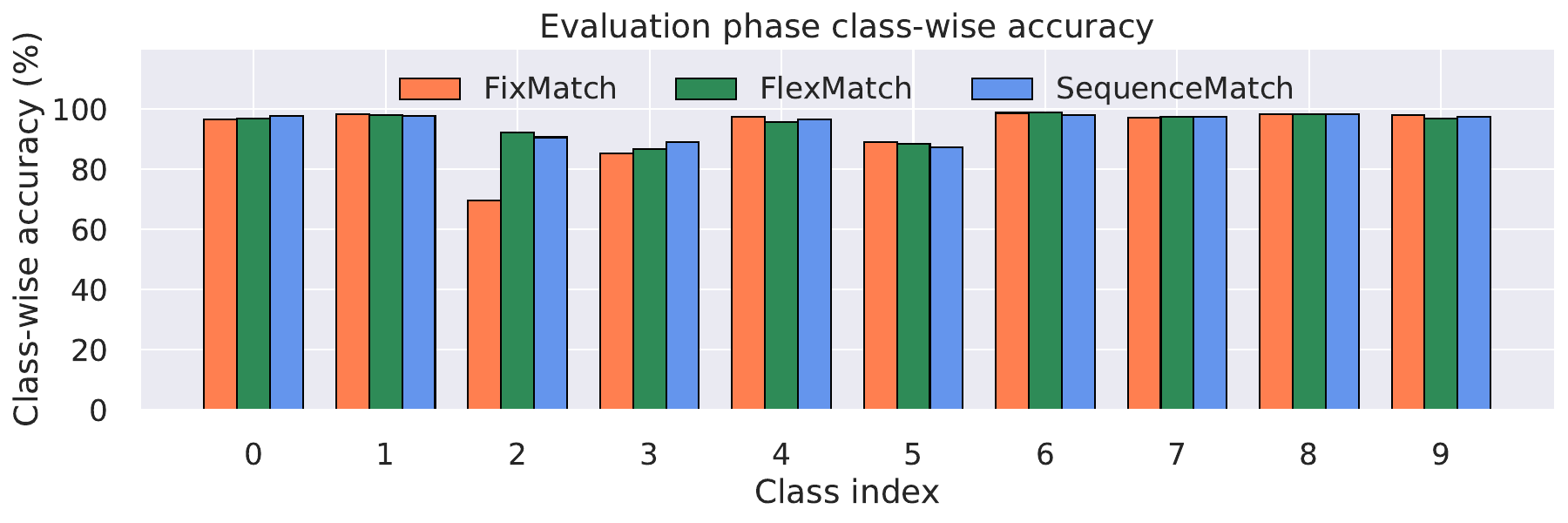}}
    \caption{Class-wise accuracy comparison on CIFAR-10 40-label split at the best iteration of FixMatch, FlexMatch and SequenceMatch.}
    \label{fig:classwise}
\end{figure}

We report a detailed comparison for class-wise accuracy in Table \ref{table:classwise}.
Our proposed SequenceMatch not only retains a high accuracy in easy-to-learn classes but also improves the accuracy of hard-to-learn classes.
The final class-wise accuracy of SequenceMatch is balanced between classes, including hard-to-learn classes (\textit{e.g.} class 2, 3).
Especially for the evaluation phase, the performance of hard-to-learn classes surpasses FixMatch by a large margin.

\begin{table*}[!ht]
\centering
% \resizebox{\textwidth}{!}{%
\caption{Class-wise accuracy comparison on CIFAR-10 40-label split.}
\label{table:classwise}
\begin{sc}
\begin{tabular}{@{}l|cccccccccc@{}}
\toprule \toprule
Class Number    & 0 & 1 & 2 & 3 & 4 & 5 & 6 & 7 & 8 & 9\\ 
\midrule \midrule
FixMatch        & 0.964 & 0.982 & 0.697 & 0.852 & \textbf{0.974} & 0.890 & 0.987 & 0.970 & 0.982 & \textbf{0.981}\\
FlexMatch       & 0.967 & 0.980 & 0.921 & 0.866 & 0.957 & 0.883 & \textbf{0.988} & 0.975 & 0.982 & 0.968\\
FreeMatch       & 0.962 & 0.984 & \textbf{0.923} & 0.874 & 0.963 & \textbf{0.894} & 0.979 & \textbf{0.977} & 0.980 & 0.976\\
\rowcolor{LightGreen}\textbf{SequenceMatch}   & \textbf{0.977} & \textbf{0.984} & 0.922 & \textbf{0.890} & 0.966 & 0.889 & 0.981 & 0.974 & \textbf{0.985} & 0.980\\ 
\bottomrule \bottomrule
\end{tabular}%
\end{sc}
% }
\end{table*}

The class-wise accuracy from the training phase, as shown in Figure \ref{fig:classwise}, also indicates that SequenceMatch can help reduce the confirmation bias issue.
It can be seen that the training phase accuracy of SequenceMatch is not only higher than FixMatch and FlexMatch but also balanced between classes.
SequenceMatch prevents the trained model from overfitting toward easy-to-learn classes.

\begin{figure}[ht]
    \centering
        \includegraphics[width=0.5\linewidth]{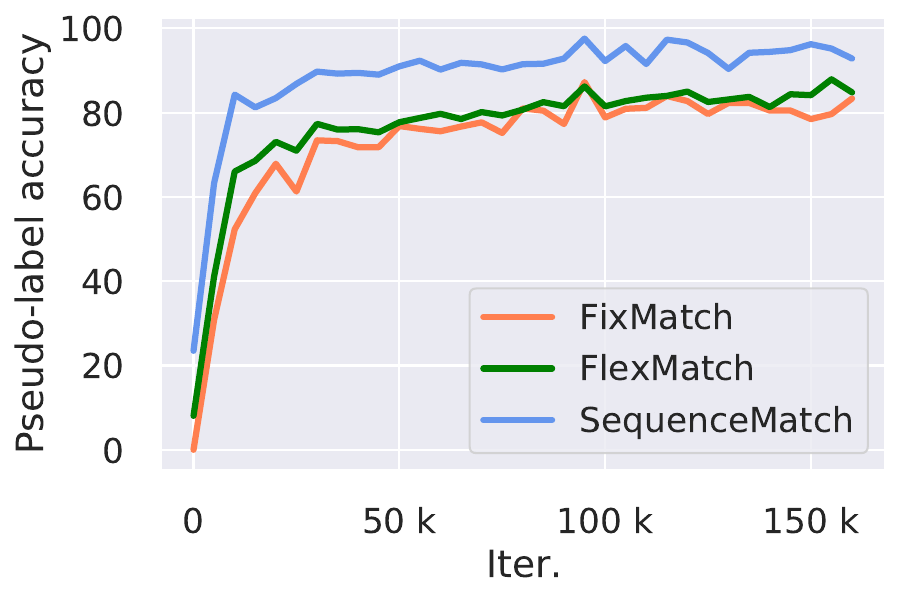}
        \caption{Pseudo-label accuracy on CIFAR-10-40.}
        \label{fig:pseudo-acc}
\end{figure}

Figure \ref{fig:pseudo-acc} shows the accuracy of the pseudo-label during training on the CIFAR-10 40-label split.
We can see that SequenceMatch mitigates the overfitting and overconfidence issues, therefore achieving a higher pseudo-label accuracy.

\begin{figure}[ht!]
    \centering
    \subfloat[Mask ratio\label{fig:mask}]
    {\includegraphics[width=.49\linewidth]{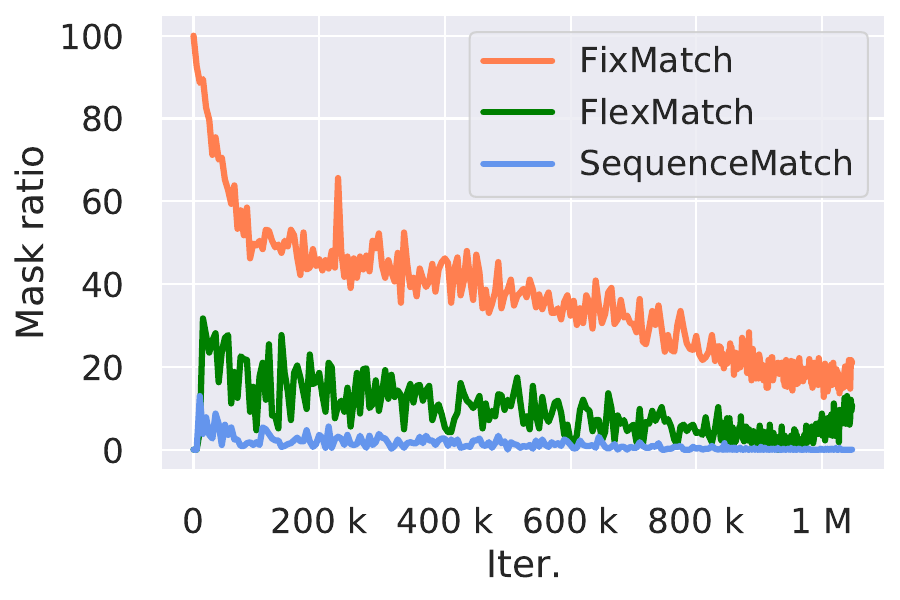}}%
    \subfloat[Data utilization ratio\label{fig:utilization}]
    {\includegraphics[width=.49\linewidth]{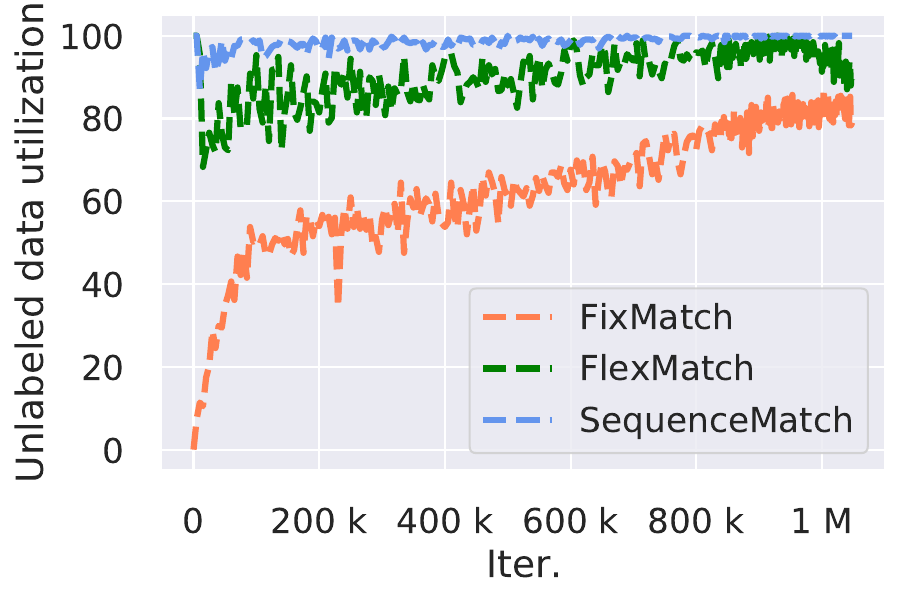}}
    \caption{Unlabeled data utilization and mask ratio on CIFAR-100 dataset with 400-label split.}
\end{figure}

\subsubsection*{Data utilization and mask ratio}
We present the unlabeled data utilization and mask ratio of FixMatch, FlexMatch, and SequenceMatch on the CIFAR-100 dataset with a 400-label split in Figure \ref{fig:mask}, \ref{fig:utilization}.
SequenceMatch significantly reduces the mask-out data ratio and is stable throughout the training process.
It also can be seen that the mask ratio of SequenceMatch fluctuates less than that of FixMatch and FlexMatch.
Furthermore, the data utilization ratio of SequenceMatch surpasses that of FixMatch and FlexMatch by a large margin.

\subsection{ImageNet}
We further evaluate SequenceMatch on the ImageNet \cite{Deng2009ImageNetAL} dataset to verify the performance on the large and complex dataset.
We compare the proposed SequenceMatch with FixMatch, FlexMatch, CoMatch, and SimMatch.
All of the models are trained on 100k of the training data as labeled.
% 100000k 
The rest of the data is treated as unlabeled data.
Furthermore, because the ImageNet dataset is large and complex, we set the $\tau$ threshold to 0.7 to improve the capture of samples with the correct pseudo-label.
The batch size is set to 128 and the weight decay is set to 0.0003.

\begin{table}[!ht]
\centering
    \caption{Error rate results on ImageNet.}
    \label{table:results2}
\begin{sc}
\resizebox{0.9\linewidth}{!}{%
    \begin{tabular}{@{}l|c|c|c|c@{}}
    \toprule \toprule
    \multirow{2}{*}{Method}                  & Top-1 & Top-5 & Top-1 & Top-5\\ 
    \cmidrule{2-5}
    & \multicolumn{2}{c|}{100k} & \multicolumn{2}{c}{10\%}\\
    \midrule \midrule
    FixMatch \cite{sohn2020fixmatch}                    & 43.66 & 21.80 & 28.50 & 10.90  \\
    FlexMatch \cite{zhang2021flexmatch}                 & 41.85 & 19.48 & - & - \\
    CoMatch \cite{li2021comatch}                        & 42.17 & 19.64 & 26.30 & 8.60 \\
    SimMatch \cite{zheng2022simmatch}                   & - & - & 25.60 & 8.40 \\
    FreeMatch \cite{wang2022freematch}                  & 40.57 & 18.77 & - & - \\
    \rowcolor{LightGreen}\textbf{SequenceMatch}           & \textbf{38.46} & \textbf{17.38} & \textbf{25.20} & \textbf{8.10} \\ \bottomrule \bottomrule
    \end{tabular}%
}
\end{sc}
\end{table}

As reported in Table \ref{table:results2}, SequenceMatch outperforms FlexMatch with 38.46\% and 17.38\% for top-1 and top-5 error rates, respectively.
The top-1 error rate result is 3.39\% lower than FlexMatch and 5.20\% lower than FixMatch.
This result strongly indicates that when the task is complicated and the dataset is imbalanced (in the ImageNet dataset, the number of images per class ranges between 732 and 1300), our proposed SequenceMatch can help boost the performance.
We also compare SequenceMatch with CoMatch and SimMatch using their source code on 10\% labeled data.
SequenceMatch outperforms FixMatch with and without using self-supervised pre-trained weights.
Compared with CoMatch and SimMatch, SequenceMatch achieves higher performance while having a fewer number of parameters.

\subsection{Imbalance dataset}
\begin{figure}[!ht]
    \centering
    \subfloat[vs FixMatch\label{fig:fdsvhn}]
    {\includegraphics[width=.49\linewidth]{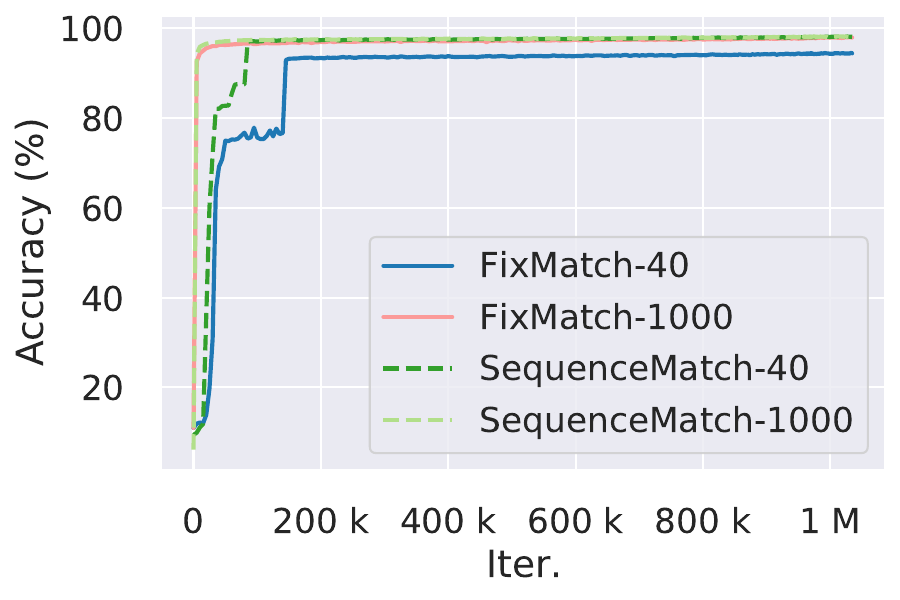}}%
    \subfloat[vs FlexMatch\label{fig:fldsvhn}]
    {\includegraphics[width=.49\linewidth]{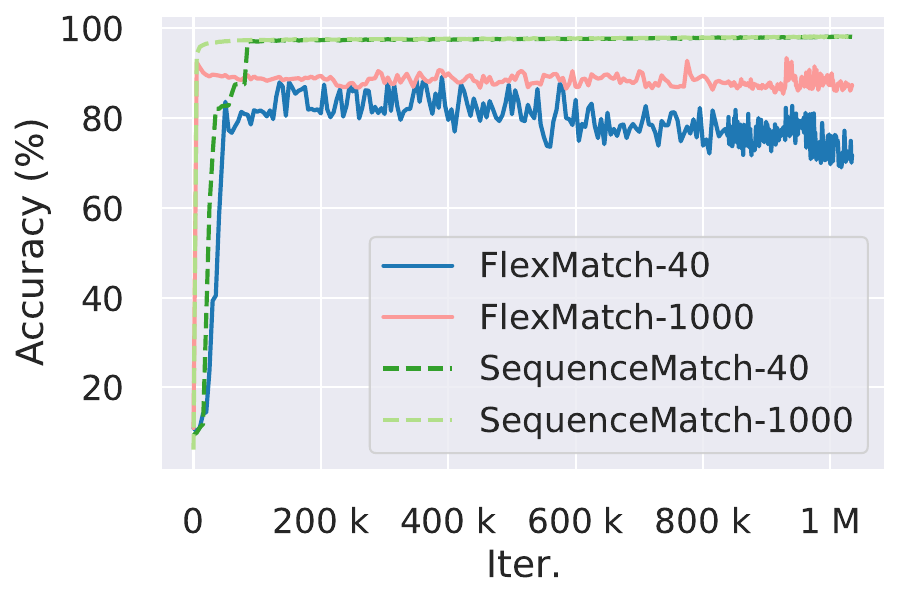}}
    \caption{Accuracy comparison of Figure \ref{fig:fdsvhn}: FixMatch vs SequenceMatch and Figure \ref{fig:fldsvhn}: Flexmatch vs SequenceMatch for first 150k iterations on SVHN dataset with 40-label and 1000-label}
    \label{fig:imb:svhn}
\end{figure}

\begin{figure}[!ht]
    \centering
    \subfloat[Top-1 accuracy\label{fig:imagenetacc}]
    {\includegraphics[width=.49\linewidth]{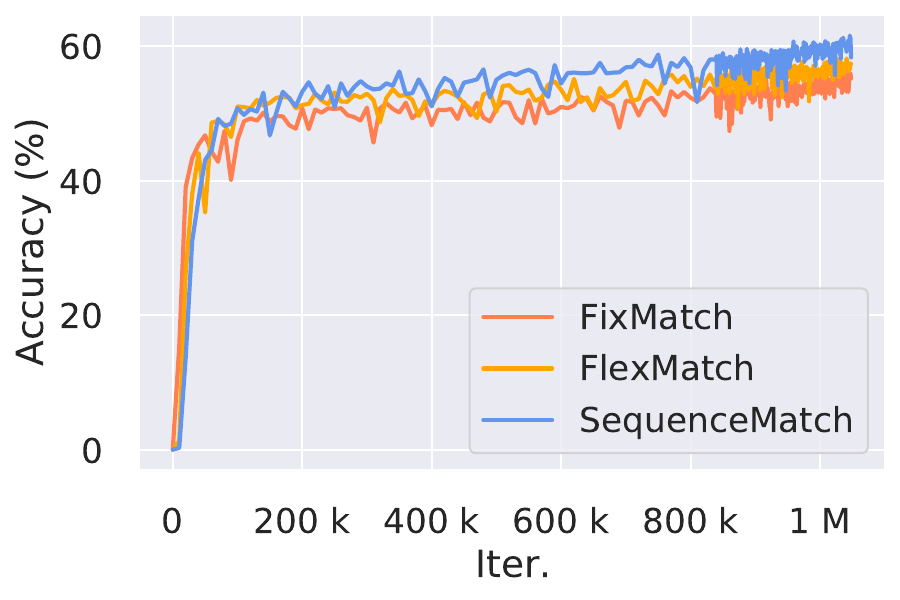}}%
    \subfloat[Eval loss\label{fig:imagenetloss}]
    {\includegraphics[width=.49\linewidth]{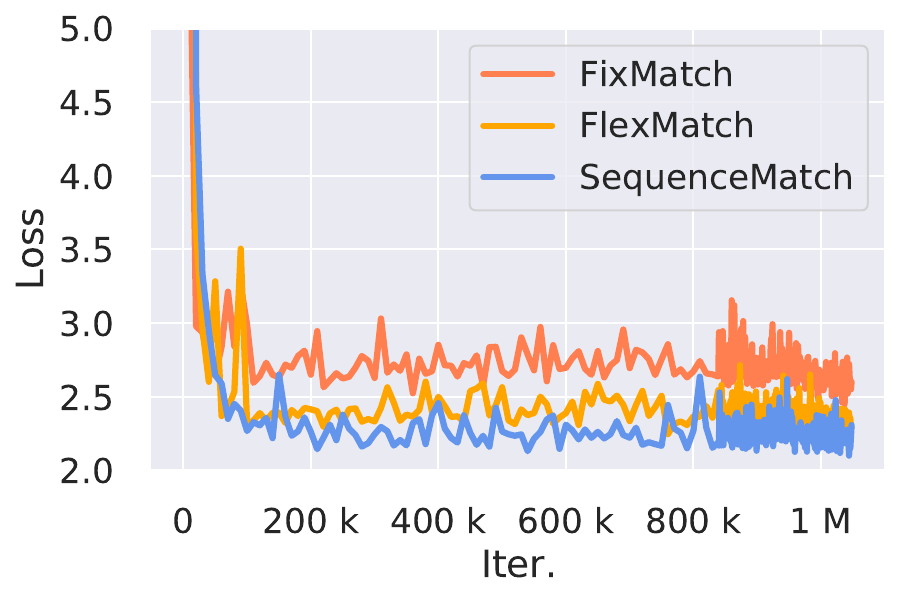}}
    \caption{Accuracy and loss comparison of Fixmatch, FlexMatch, and SequenceMatch on Imagenet dataset.}
    \label{fig:imb:imagenet}
\end{figure}

In Figure \ref{fig:imb:svhn} and \ref{fig:imb:imagenet}, we show the performance of FixMatch, FlexMatch, and SequenceMatch on the SVHN and ImageNet datasets.
For instance, our proposed SequenceMatch results show superiority over FixMatch and FlexMatch when dealing with imbalanced data problems such as SVHN and ImageNet datasets.
According to Table \ref{table:results1}, Table \ref{table:results2}, our results are identical to FlexMatch results; however, FlexMatch fails on the SVHN dataset since CPL may generate low final thresholds for the tail classes that allow noisy pseudo-labeled samples to be trusted and learned.
SequenceMatch solves this problem by maintaining the consistency of the model throughout the training process and mitigating the overfitting issue.
Furthermore, SequenceMatch results on the ImageNet dataset outperform FixMatch and FlexMatch without additional modules.

\subsection{Calibration of SSL}
\begin{figure*}[!ht]
    \centering
\resizebox{0.6\linewidth}{!}{%
    \subfloat[FixMatch\label{fig:ece1}]
    {\includegraphics[width=.33\linewidth]{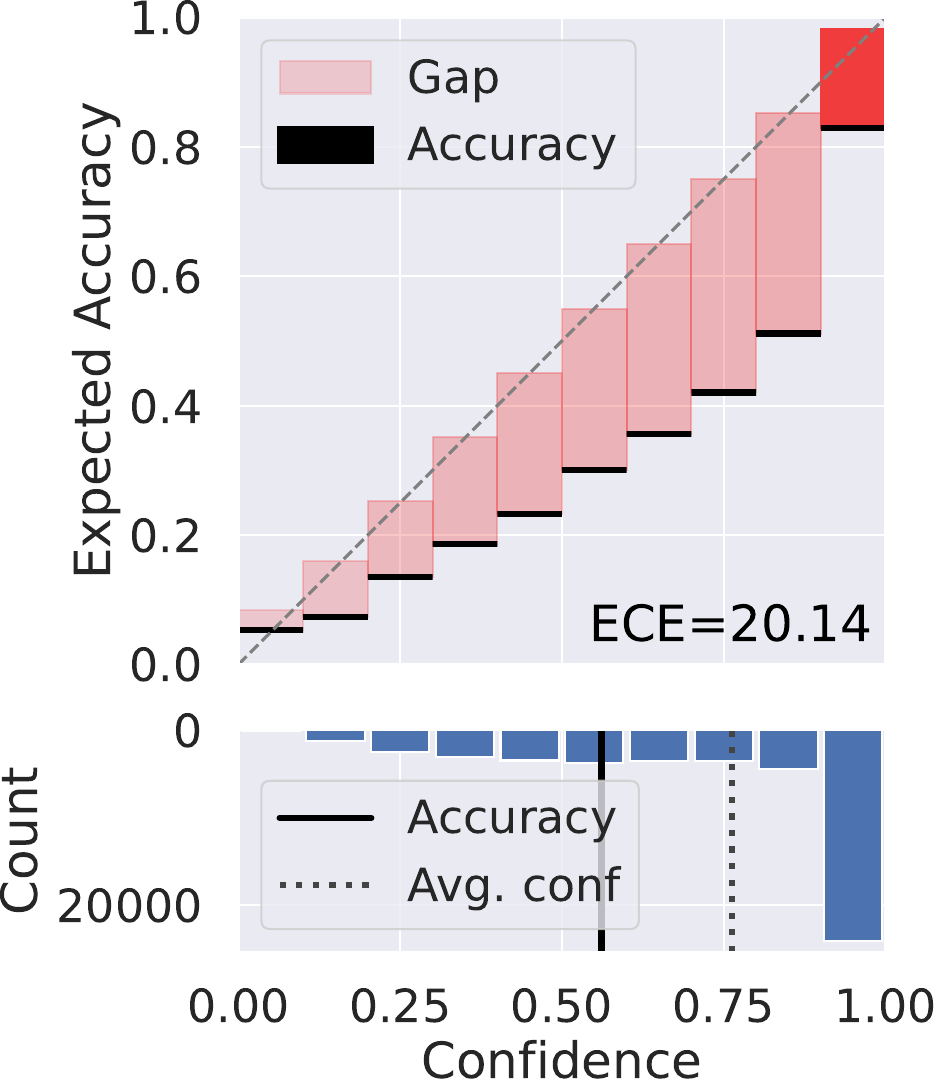}}%
    \subfloat[FlexMatch\label{fig:ece2}]
    {\includegraphics[width=.33\linewidth]{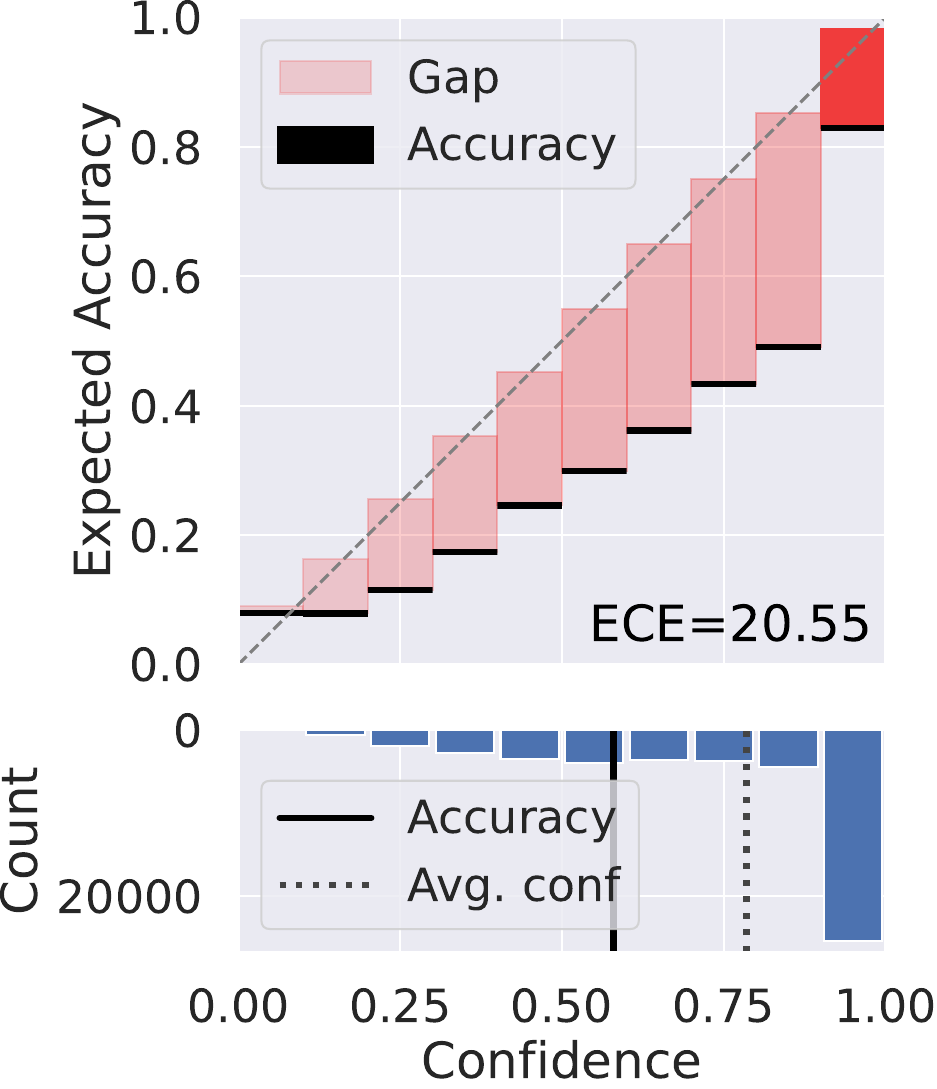}}%
    \subfloat[SequenceMatch\label{fig:ece3}]
    {\includegraphics[width=.33\linewidth]{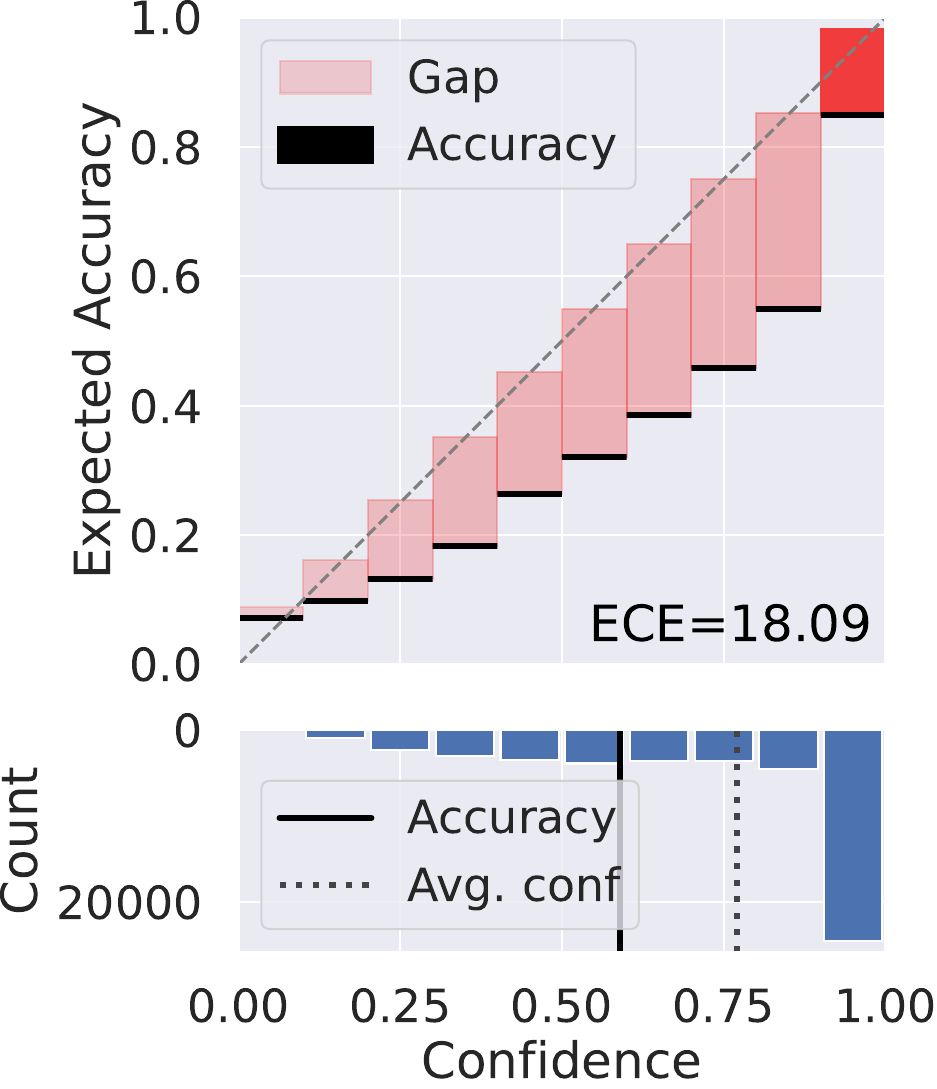}}%
}
    \caption{Reliability diagrams (top) and confidence histograms (bottom) for ImageNet dataset.}
    \label{fig:ece}
\end{figure*}

\cite{chen2022semi} suggests addressing confirmation bias from the calibration perspective.
We measure the calibration of FixMatch, FlexMatch, and SequenceMatch trained on the ImageNet dataset with 100k labels \footnote{https://github.com/hollance/reliability-diagrams}.
Several common calibration indicators are used: Expected Calibration Error (ECE), confidence histogram, and reliability diagram.
As shown in Figure \ref{fig:ece}, even though FlexMatch has higher accuracy than FixMatch, its ECE value of $20.55$ is larger than that of FixMatch, at $20.14$, indicating poorer probability estimation.
On the other hand, SequenceMatch achieves both higher accuracy and a lower ECE value of $18.09$, which proves that it can reduce the confirmation bias and produce a well-calibrated model.

\section{Related Work}
% \subsection{Semi-supervised Learning}
Self-training is a concept that has been around for decades \cite{mclachlan1975iterative,scudder1965probability}.
Self-training (i.e., utilizing a prediction distribution to generate pseudo-labels for unlabeled data) has been employed in a variety of areas, including natural language processing \cite{mcclosky2006effective}, object recognition \cite{Rosenberg2005SemiSupervisedSO}, image classification \cite{lee2013pseudo,Xie2020SelfTrainingWN}, domain adaption \cite{zou2018unsupervised}, etc.
Pseudo-labeling \cite{lee2013pseudo} is a pioneering SSL method that uses "hard" artificial labels converted from model predictions.
Pseudo-labeling is frequently used together with confidence-based thresholding, which keeps unlabeled samples only when predictions are sufficiently confident \cite{Rosenberg2005SemiSupervisedSO,xie2020unsupervised,sohn2020fixmatch,zhang2021flexmatch}.

\cite{Bachman2014LearningWP} introduced consistency regularization, which was later popularized \cite{laine2016temporal,Sajjadi2016RegularizationWS}.
Consistency regularization utilizes unlabeled data by relying on the assumption that the model should output similar predictions when perturbed versions of the same image are fed.
Data augmentation \cite{french2017self}, stochastic regularization \cite{laine2016temporal,sajjadi2016regularization}, and adversarial perturbations \cite{Miyato2019VirtualAT} have all been used to generate random perturbations.
It has recently been demonstrated that applying significant data augmentation can improve outcomes \cite{xie2020unsupervised}.
% These strategies are combined into a self-training framework by Noisy Student \cite{Xie2020SelfTrainingWN}, which has an exhibited outstanding performance on ImageNet with the additional massive amount of unlabeled data.

FixMatch \cite{sohn2020fixmatch} proposed a combination of both pseudo-labeling and consistency regularization methods for SSL.
FixMatch's thresholded pseudo-labeling produces a sharpening-like effect that encourages the model to deliver high-confidence predictions.
FixMatch could be considered a combination version of UDA and ReMixMatch, in which two common strategies (pseudo-labeling and consistency regularization) are integrated while many components are removed (sharpening, training signal annealing from UDA, distribution alignment, and the rotation loss from ReMixMatch, etc.).
FlexMatch \cite{zhang2021flexmatch} introduces a Curriculum Pseudo Labeling (CPL) method, which enables conventional SSL to train with a dynamic threshold for each class.
CPL can be considered a dynamic thresholding approach since it dynamically adjusts the threshold for each class after each iteration, thus enabling higher performance for each class.
FlexMatch outperforms most state-of-the-art SSL across a wide range of datasets.

Lately, CoMatch \cite{li2021comatch} is introduced, which combines two contrastive representations on unlabeled data.
However, CoMatch is extremely sensitive to hyperparameter settings and requires a large memory bank during training to store the embedded features.
Recently work of \cite{zheng2022simmatch} considers the semantic similarity and instance similarity during training.
It shows that forcing consistency on both the semantic-level and instance-level can bring an improvement, thus achieving state-of-the-art benchmarks.

% \subsection{"Hard" and "soft" targets}
% \cite{hinton2015distilling} show that a lot of helpful information can be carried in "soft" targets that could not possibly be encoded with a single "hard" target.
% Additionally, SSL uses a small amount of labeled data; hence, training the baseline model with "hard" targets leads to severe overfitting.

% Recently, \cite{botev2022regularising} shows that leveraging multiple data augmentation with KL divergence for "soft" targets could encourage invariance and generalization for supervised learning.
% However, using "soft" targets with a high-confidence threshold in FixMatch does not show a significant difference in performance.
% Therefore, we mitigate this insufficiency in our method and propose to use both "hard" and "soft" targets with multiple data augmentations.

\section{Conclusion}
In this paper, we introduce SequenceMatch, an SSL pipeline that sequentially matches predictions to reduce the divergence between the predicted class distributions for different augmented versions of the same input.
SequenceMatch introduces a medium augmentation for unlabeled data, which helps reduce the divergence between the prediction distributions while maintaining the correct pseudo-label.
Furthermore, SequenceMatch also helps reduce the overfitting phenomenon that most SSL methods are facing.
SequenceMatch achieves state-of-the-art performance on a variety of SSL benchmarks and works well for all datasets.

%%%%%%%%% REFERENCES
{\small
\bibliographystyle{ieee_fullname}
\bibliography{egbib}
}

% APPENDIX
%%%%%%%%%%%%%%%%%%%%%%%%%%%%%%%%%%%%%%%%%%%%%%%%%%%%%%%%%%%%%%%%%%%%%%%%%%%%%%%
%%%%%%%%%%%%%%%%%%%%%%%%%%%%%%%%%%%%%%%%%%%%%%%%%%%%%%%%%%%%%%%%%%%%%%%%%%%%%%%
\clearpage
\appendix
\section*{Qualitative analysis}
\begin{figure}[!ht]
    \centering
    \subfloat[\footnotesize{FixMatch}\label{fig:tsne-fix}]
    {\includegraphics[width=.3\linewidth]{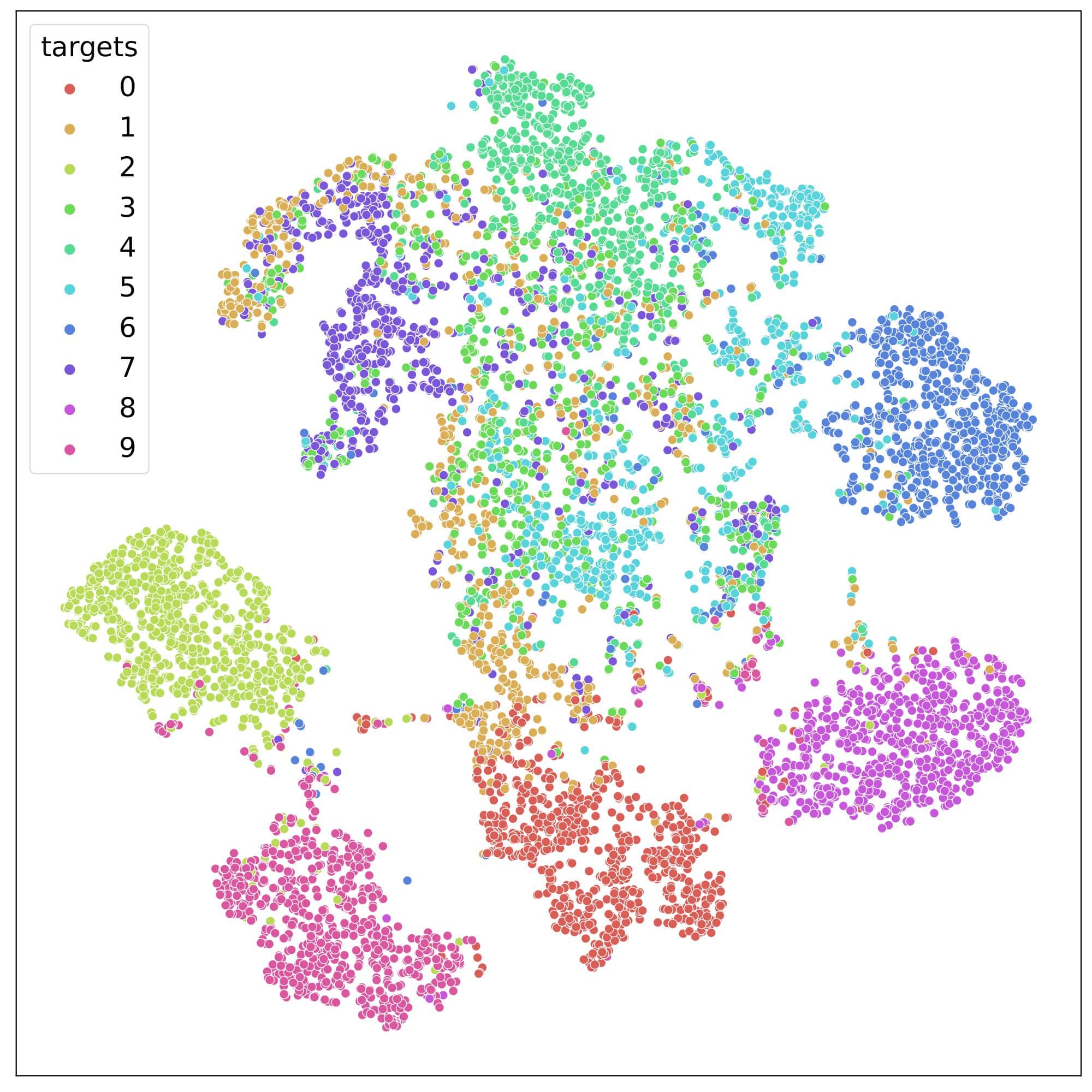}}%
    \subfloat[\footnotesize{FlexMatch}\label{fig:tsne-flex}]
    {\includegraphics[width=.3\linewidth]{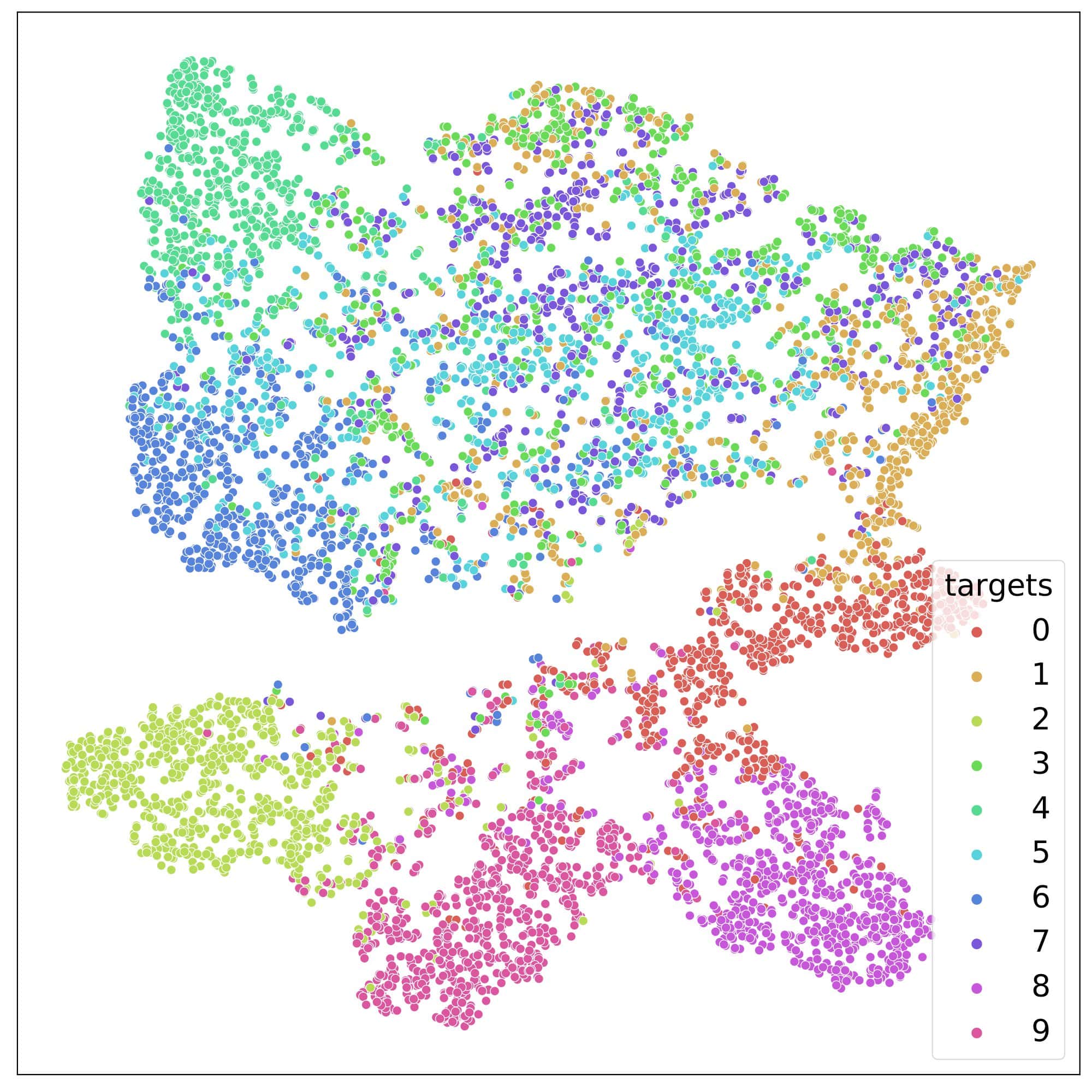}}%
    \subfloat[\footnotesize{SequenceMatch}\label{fig:tsne-seq}]
    {\includegraphics[width=.3\linewidth]{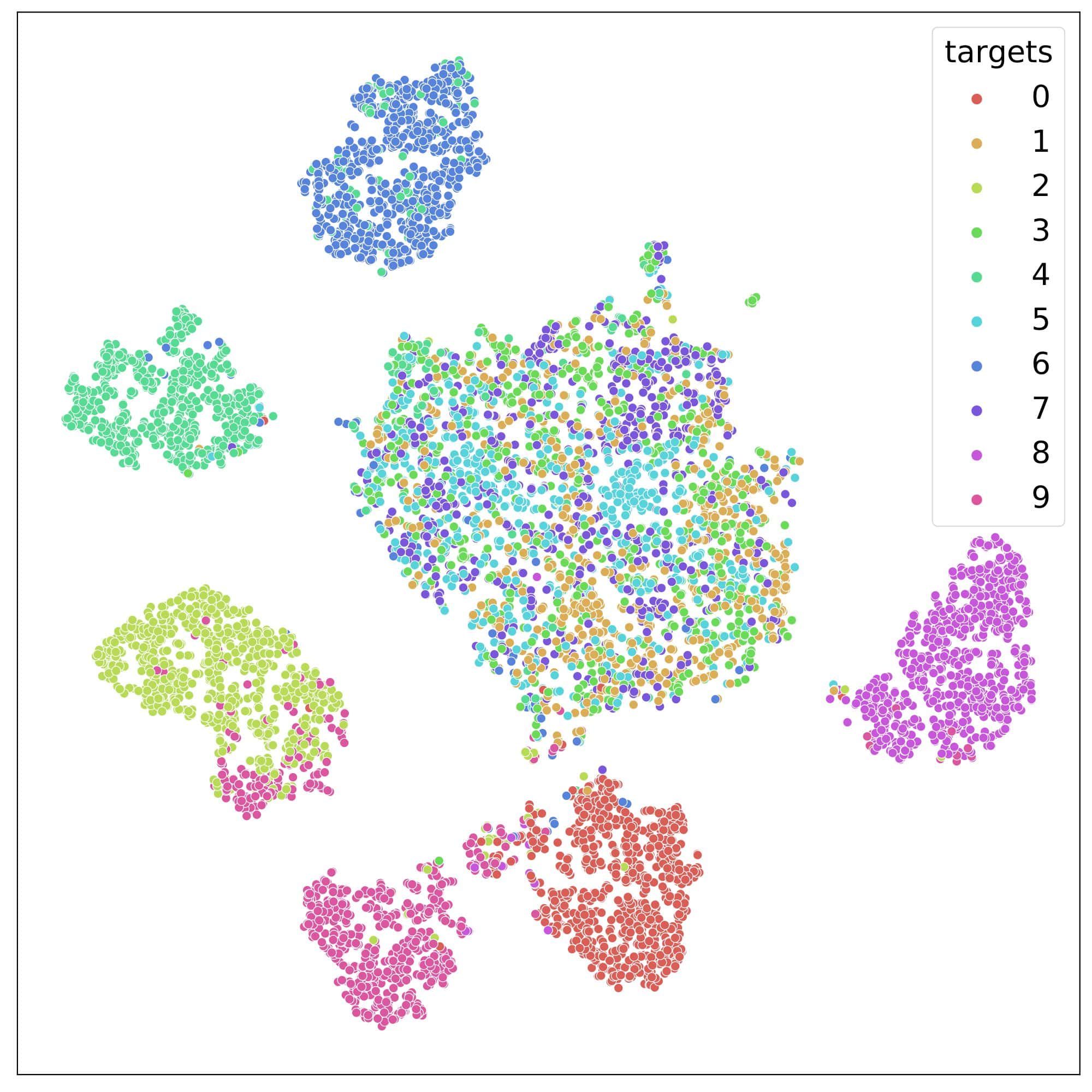}}%
    \caption{T-SNE visualization on STL-10 dataset with 40 labels.}
\end{figure}

We present the T-SNE visualization of features on the STL-10 test dataset with a 40-label split in Figure \ref{fig:tsne-fix},\ref{fig:tsne-flex},\ref{fig:tsne-seq}.
The visualization are using trained models from FixMatch, FlexMatch, and SequenceMatch.
SequenceMatch shows better feature space than FixMatch and FlexMatch with less confusing clusters.

\begin{figure}[!ht]
    \centering
    \subfloat[\footnotesize{FixMatch}\label{fig:tsne-fix-svhn}]
    {\includegraphics[width=.3\linewidth]{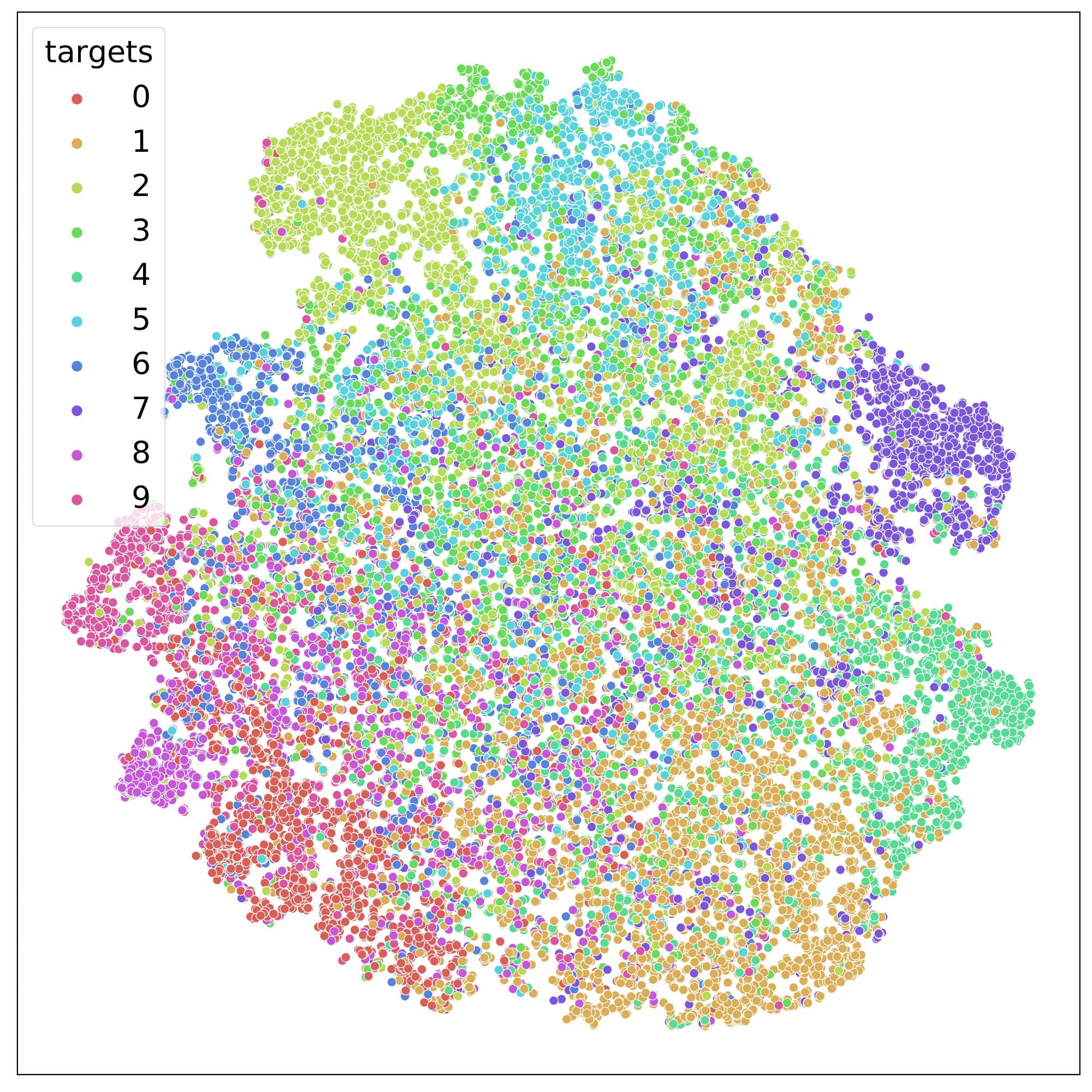}}%
    \subfloat[\footnotesize{FlexMatch}\label{fig:tsne-flex-svhn}]
    {\includegraphics[width=.3\linewidth]{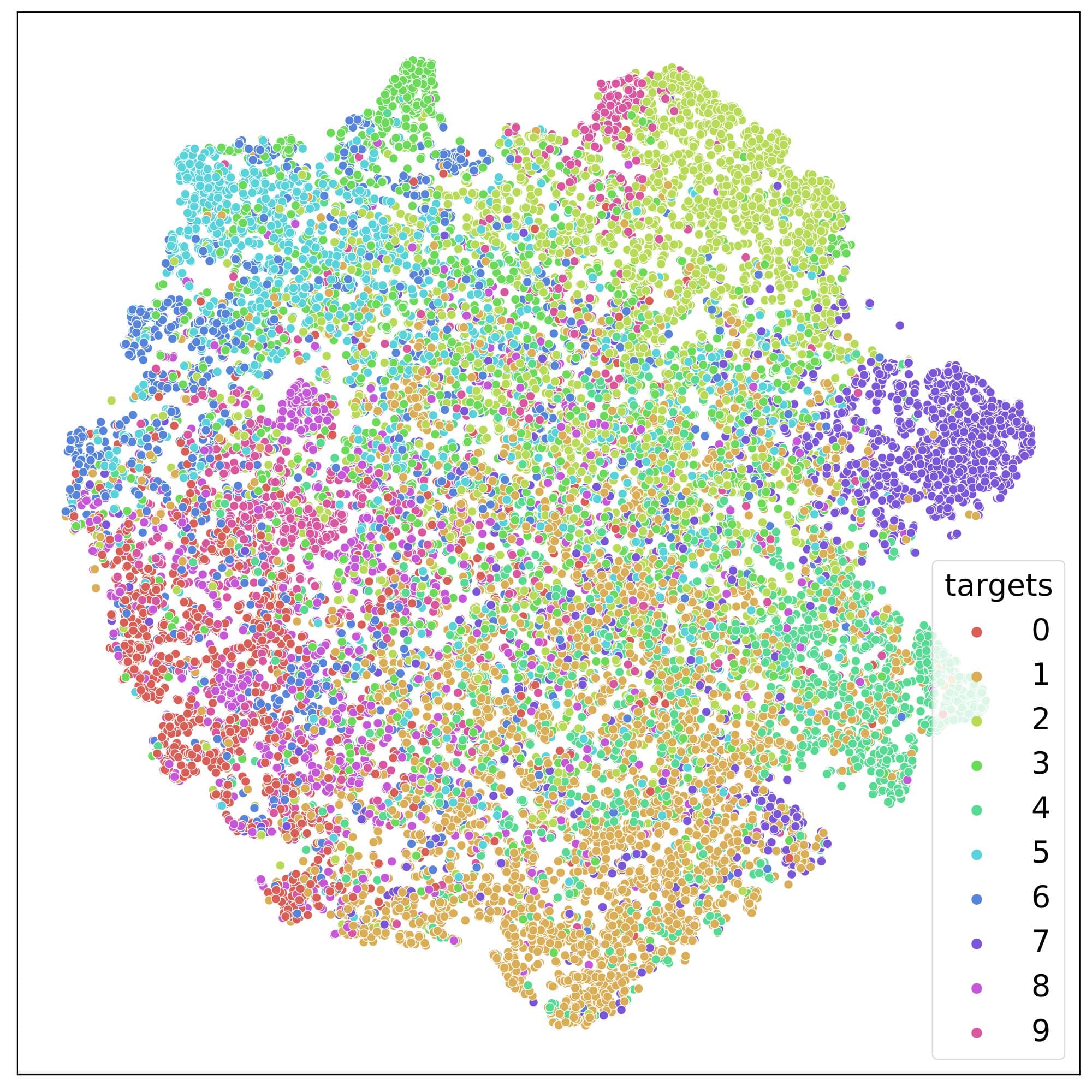}}%
    \subfloat[\footnotesize{SequenceMatch}\label{fig:tsne-seq-svhn}]
    {\includegraphics[width=.3\linewidth]{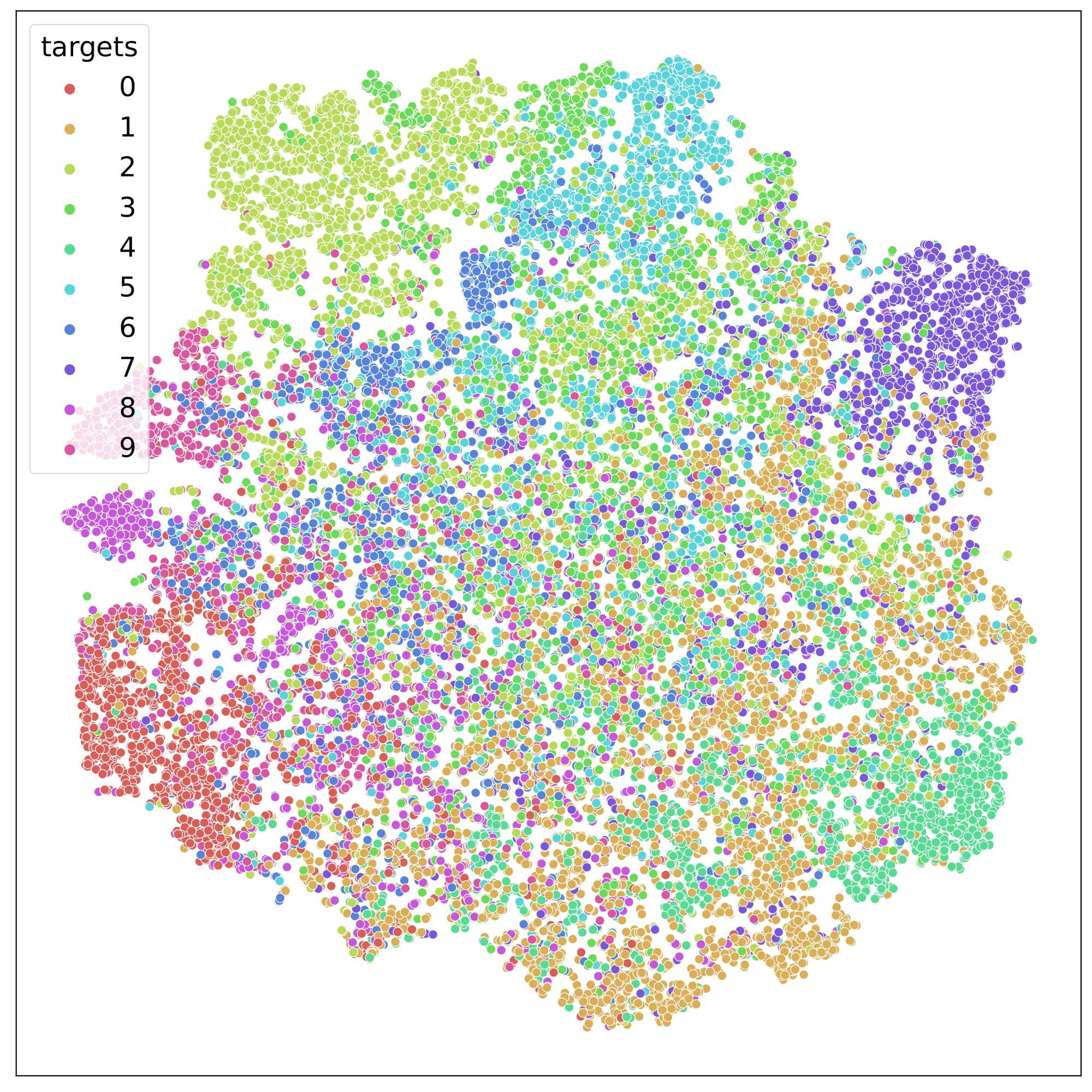}}%
    \caption{T-SNE visualization on SVHN dataset with 40 labels.}
\end{figure}

\begin{figure}[!ht]
    \centering
    \subfloat[\footnotesize{FixMatch}\label{fig:tsne-fix-cifar}]
    {\includegraphics[width=.3\linewidth]{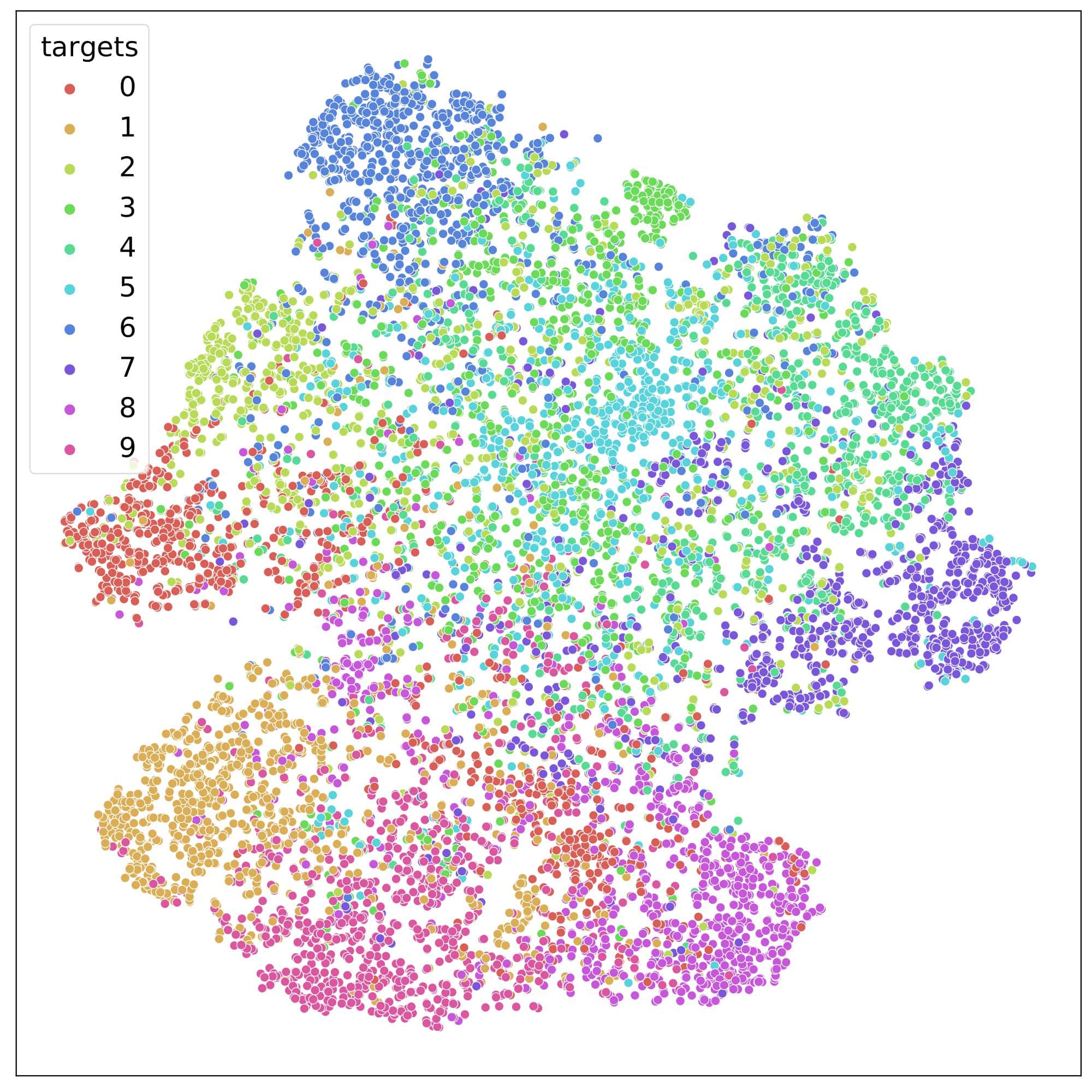}}%
    \subfloat[\footnotesize{FlexMatch}\label{fig:tsne-flex-cifar}]
    {\includegraphics[width=.3\linewidth]{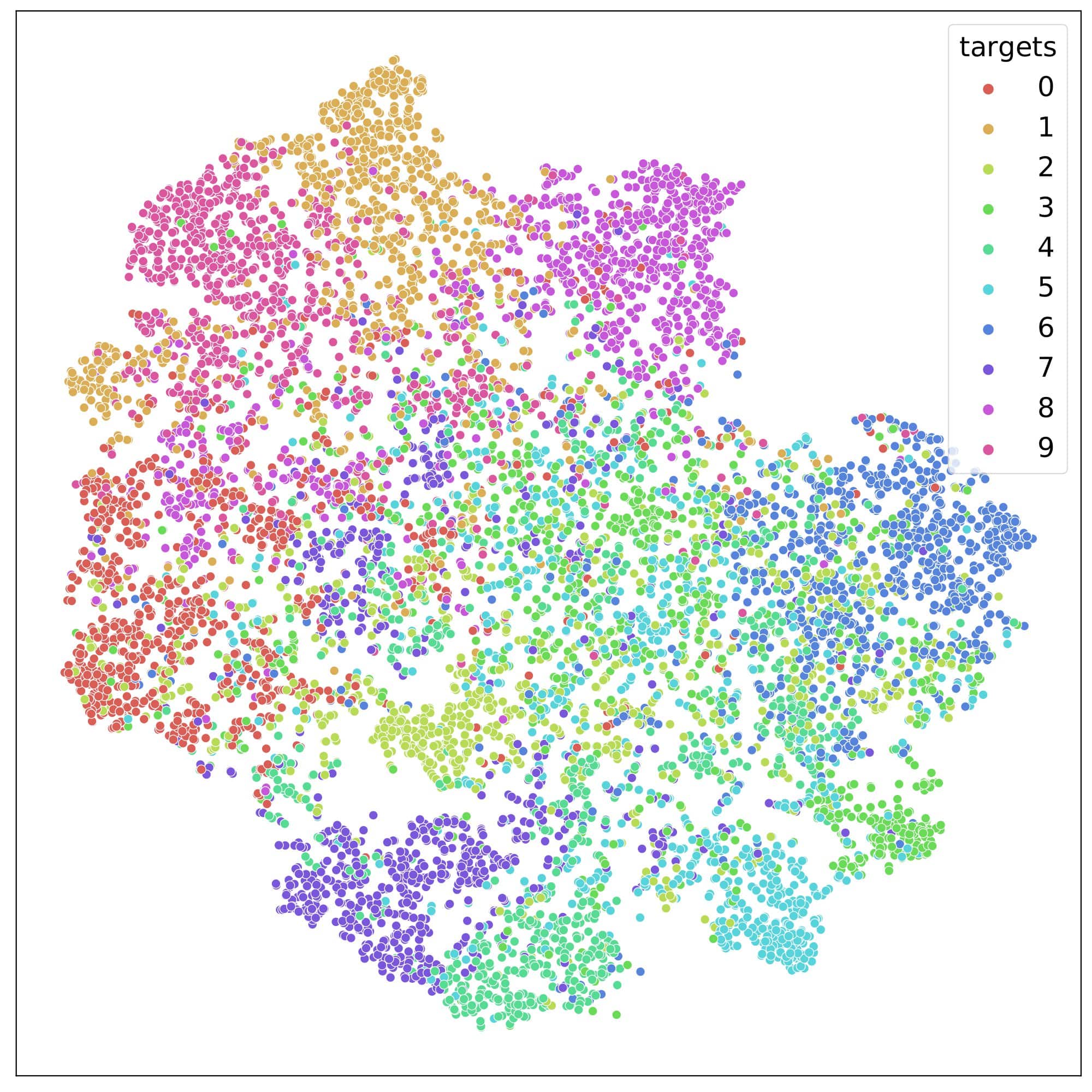}}%
    \subfloat[\footnotesize{SequenceMatch}\label{fig:tsne-seq-cifar}]
    {\includegraphics[width=.3\linewidth]{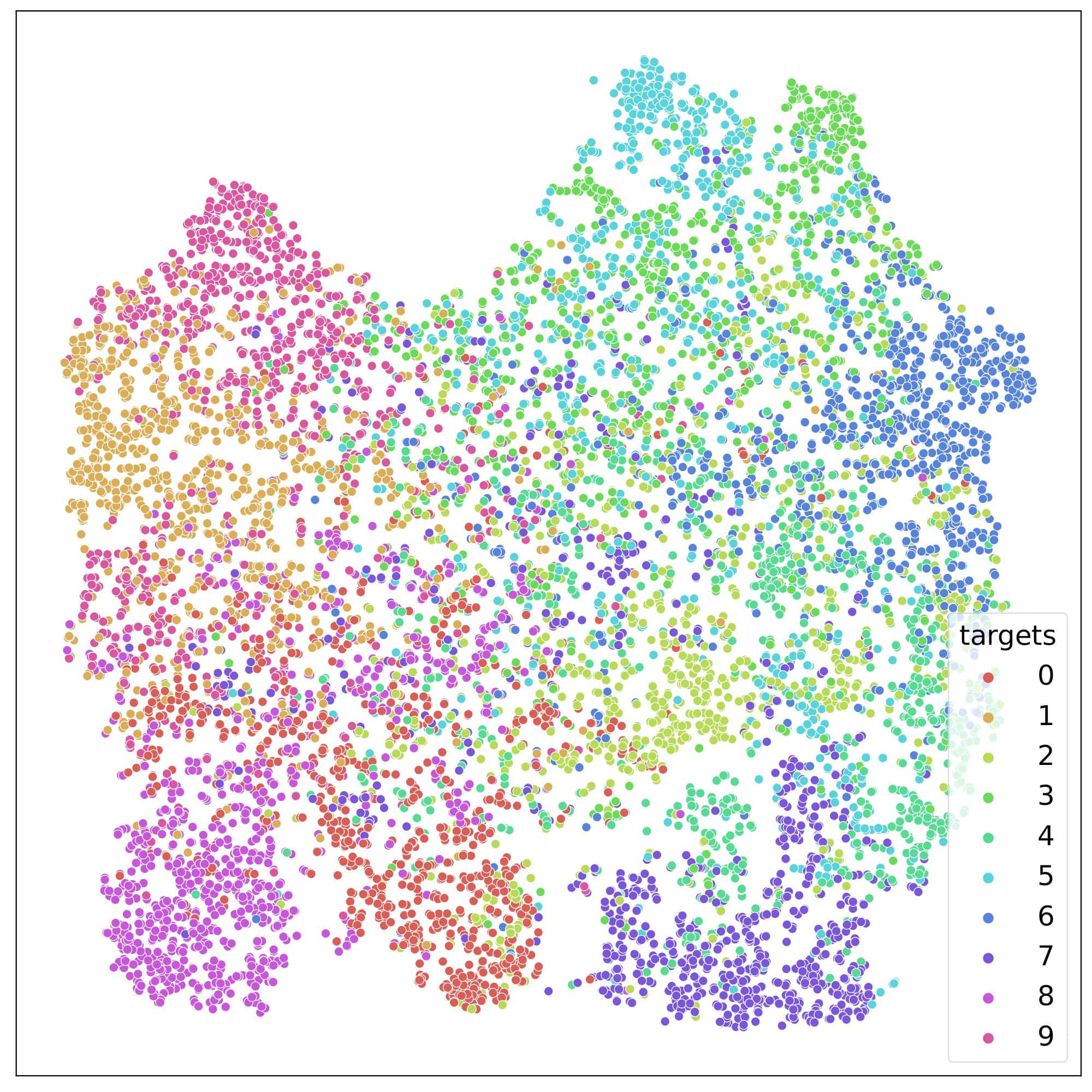}}%
    \caption{T-SNE visualization on CIFAR-10 dataset with 40 labels.}
\end{figure}

We also visualize the T-SNE visualization of features on the SVHN test dataset and CIFAR-10 test dataset with a 40-label split in Figure \ref{fig:tsne-fix-svhn},\ref{fig:tsne-flex-svhn},\ref{fig:tsne-seq-svhn} and Figure \ref{fig:tsne-fix-cifar},\ref{fig:tsne-flex-cifar},\ref{fig:tsne-seq-cifar}, respectively.

\section*{Hyperparameters setting}
\label{sec:app}
For reproduction, we show the detailed hyperparameter setting for each method in Table \ref{table:hyperparameter1} and Table \ref{table:hyperparameter2}, for algorithm-dependent and algorithm-independent hyperparameters, respectively.

\begin{table*}[ht!]
\centering
% \resizebox{\linewidth}{!}{
    \caption{Algorithm dependent parameters. \textbf{'F-Match'} indicates FixMatch, FlexMatch, and FreeMatch.}
\begin{small}
\begin{sc}
    \begin{tabular}{@{}l|cccc@{}}
    \toprule \toprule
    Algorithm   & UDA   & SequenceMatch  & F-Match          \\ \midrule
    Unlabeled Data to Labeled Data Ratio   & \multirow{2}{*}{7}     & \multirow{2}{*}{7}    & \multirow{2}{*}{7} \\
    (CIFAR-10/100, STL-10, SVHN)   &       &        &       \\ \midrule
    Unlabeled Data to Labeled Data Ratio     & \multirow{2}{*}{-}         & \multirow{2}{*}{1}         & \multirow{2}{*}{1} \\ 
    (ImageNet)          &           &           & \\ \midrule
    Pre-defined Threshold                               & \multirow{2}{*}{0.8}  & \multirow{2}{*}{0.95}       & \multirow{2}{*}{0.95}                             \\
    (CIFAR-10/100, STL-10, SVHN)                        &           &           & \\ \midrule
    Pre-defined Threshold (ImageNet)                    & -         & 0.7       & 0.7 \\ \midrule 
    Temperature                                         & 0.4       & 0.5       & - \\ \bottomrule \bottomrule
    \end{tabular}%
\end{sc}
\end{small}
% }
    \label{table:hyperparameter1}
\end{table*}

\begin{table*}[ht!]
\centering
% \resizebox{\linewidth}{!}{
    \caption{Algorithm independent parameters.}
    \label{table:hyperparameter2}
\begin{small}
\begin{sc}
    \begin{tabular}{@{}l|ccccc}
    \toprule \toprule
    Dataset         & CIFAR-10      & CIFAR-100     & STL-10    & SVHN      & ImageNet  \\ \midrule
    Model           & WRN-28-2      & WRN-28-8      & WRN-37-2  & WRN-28-2  & ResNet-50 \\ \midrule
    Weight Decay    & 5e-4          & 1e-3          & 5e-4      & 5e-4      & 3e-4      \\ \midrule
    Batch Size      &  \multicolumn{4}{c}{64}                               & 128       \\ \midrule
    Learning Rate   & \multicolumn{5}{c}{0.03}                                          \\ \midrule
    SGD Momentum    & \multicolumn{5}{c}{0.9}                                           \\ \midrule
    EMA Momentum    & \multicolumn{5}{c}{0.999}                                         \\ \midrule
    Unsupervised Loss Weight & \multicolumn{5}{c}{1}                                    \\ \bottomrule \bottomrule
    \end{tabular}%
\end{sc}
\end{small}
    % }
\end{table*}

\section*{Ablation study on KL loss}
The additional medium augmentation requires the model to adjust to the new distribution.
This is not the case with strong augmentation since the strongly augmented sample is heavily distorted, making it impossible to retrieve relevant information.
As a consequence, SequenceMatch without and with KL loss obtains 5.01\% and \textbf{4.80\%} for CIFAR-10-40, respectively.
Furthermore, after 150k iterations, the pseudo-label accuracy is 81.1\%, 81.50\%, and \textbf{83.20\%} for FixMatch, FlexMatch, and SequenceMatch, respectively.
SequenceMatch enhances the pseudo-label accuracy while improving the hard-to-learn class-wise accuracy simultaneously.
This clearly demonstrates that employing medium augmentation and KL loss can reduce divergence and eliminate a confirmation bias.

\begin{table*}[ht!]
\begin{minipage}[h]{0.45\textwidth}
\centering
\caption{KL loss and results on CIFAR-10 with 40-label split.}
\label{table:kl}
\begin{small}
\begin{sc}
    \begin{tabular}{l|cc}
    \toprule \toprule
                                    &   with KL loss    &   Top-1\\
    \midrule \midrule
    \multirow{2}{*}{SequenceMatch}  &   $\times$        &   5.01\\
                                    &   $\surd$         &   4.80\\
    \bottomrule \bottomrule
    \end{tabular}%
\end{sc}
\end{small}
\end{minipage}
\begin{minipage}[h]{0.55\linewidth}
\centering
\caption{Augmentation results on CIFAR-10 with 40-label split.}
    \label{table:augmentation}
\resizebox{\linewidth}{!}{
\begin{small}
\begin{sc}
    \begin{tabular}{l|c}
    \toprule \toprule
    Method                                                  &   Top-1\\
    \midrule \midrule
    \rowcolor{LightGreen}Weak augmentation + 1 random strong augmentation        &   \textbf{4.80}\\
    Weak augmentation + 2 random strong augmentation        &   4.91\\
    MoCo augmentation \cite{he2020momentum}                                       &   5.97\\
    CTAugment \cite{Berthelot2020ReMixMatchSL}                                               &   4.85\\
    SimCLR augmentation \cite{chen2020simple}                                          &   5.32\\    
    \bottomrule \bottomrule
    \end{tabular}%
\end{sc}
\end{small}
}
\end{minipage}
\end{table*}

\begin{table*}[ht!]
\centering
\caption{Mean error rates of last 20 checkpoints of all methods. 
There are 1000 iterations between every two checkpoints.}
    \label{table:mer}
\resizebox{\textwidth}{!}{
\begin{small}
\begin{sc}
    \begin{tabular}{@{}l|ccc|ccc|ccc|cc}
    \toprule \toprule
    Dataset &   \multicolumn{3}{c|}{CIFAR-10}    & \multicolumn{3}{c|}{CIFAR-100} & \multicolumn{3}{c|}{SVHN}    &   \multicolumn{2}{c}{STL-10}\\
    \midrule 
    \# Label &   40  &   250 &   4000    &   400 &   2500    &   10000   &   40  &   250 &   1000   &   40  &   1000\\
    \midrule \midrule
    UDA &   10.65\scriptsize{$\pm$4.97}  &   5.67\scriptsize{$\pm$0.28}  &   4.58\scriptsize{$\pm$0.07}  &   99.0\scriptsize{$\pm$0.0}  &   99.0\scriptsize{$\pm$0.0}  &   99.0\scriptsize{$\pm$0.0}  &   2.5\scriptsize{$\pm$0.54}  &   2.06\scriptsize{$\pm$0.02}  &   2.01\scriptsize{$\pm$0.03}  &   90.0\scriptsize{$\pm$0.0}  &   34.88\scriptsize{$\pm$38.98} \\
    MPL &   8.2\scriptsize{$\pm$1.9}  &   8.7\scriptsize{$\pm$1.22}  &   4.78\scriptsize{$\pm$0.03}  &   48.72\scriptsize{$\pm$0.46}  &   29.02\scriptsize{$\pm$0.46}  &   22.39\scriptsize{$\pm$0.38}  &   11.06\scriptsize{$\pm$6.45}  &   2.45\scriptsize{$\pm$0.08}  &   2.29\scriptsize{$\pm$0.06}  &   44.63\scriptsize{$\pm$7.16}  &   7.51\scriptsize{$\pm$0.19} \\
    MixMatch & 51.5\scriptsize{$\pm$17.51}  &   22.14\scriptsize{$\pm$2.83}  &   66.57\scriptsize{$\pm$5.38}  &   95.87\scriptsize{$\pm$0.24}  &   97.88\scriptsize{$\pm$0.36}  &   99.0\scriptsize{$\pm$0.0}  &   48.86\scriptsize{$\pm$14.71}  &   10.16\scriptsize{$\pm$2.7}  &   30.09\scriptsize{$\pm$2.42}  &   64.99\scriptsize{$\pm$2.3} &   59.5\scriptsize{$\pm$2.48} \\
    ReMixMatch & 8.5\scriptsize{$\pm$0.6}  &   6.59\scriptsize{$\pm$0.18}  &   4.97\scriptsize{$\pm$0.13}  &   42.1\scriptsize{$\pm$1.35}  &   26.19\scriptsize{$\pm$0.15}  &   \textbf{20.57\scriptsize{$\pm$0.14}}  &   21.41\scriptsize{$\pm$12.26}  &   10.69\scriptsize{$\pm$0.73}  &   11.44\scriptsize{$\pm$1.91}  &   34.12\scriptsize{$\pm$5.27}  &   6.99\scriptsize{$\pm$0.08} \\
    FixMatch & 12.85\scriptsize{$\pm$4.51}  &   5.26\scriptsize{$\pm$0.08}  &   4.43\scriptsize{$\pm$0.02}  &   48.87\scriptsize{$\pm$2.48}  &   28.84\scriptsize{$\pm$0.4}  &   22.93\scriptsize{$\pm$0.14}  &   3.5\scriptsize{$\pm$1.05}  &   2.06\scriptsize{$\pm$0.01}  &   2.11\scriptsize{$\pm$0.02}  &   46.71\scriptsize{$\pm$5.25}  &   6.14\scriptsize{$\pm$0.25} \\
    FlexMatch & 5.53\scriptsize{$\pm$0.28}  &   5.24\scriptsize{$\pm$0.08}  &   4.49\scriptsize{$\pm$0.04}  &   47.56\scriptsize{$\pm$2.68} &	27.62\scriptsize{$\pm$0.11}  &   22.88\scriptsize{$\pm$0.17}  &   18.55\scriptsize{$\pm$8.46} &	19.17\scriptsize{$\pm$4.63}  &   12.93\scriptsize{$\pm$1.85}  &   51.15\scriptsize{$\pm$15.35} &   6.34\scriptsize{$\pm$0.37} \\
    Dash & 9.96\scriptsize{$\pm$3.45}  &   5.38\scriptsize{$\pm$0.29}  &   4.6\scriptsize{$\pm$0.12}  &   50.37\scriptsize{$\pm$1.77}  &   28.61\scriptsize{$\pm$0.38}  &   22.85\scriptsize{$\pm$0.15}  &   5.39\scriptsize{$\pm$2.05}  &   2.08\scriptsize{$\pm$0.02}  &   2.16\scriptsize{$\pm$0.09}  &   44.19\scriptsize{$\pm$6.07}  & 6.61\scriptsize{$\pm$0.52} \\
    CoMatch & 7.2\scriptsize{$\pm$1.77}  &   5.64\scriptsize{$\pm$0.17}  &   4.52\scriptsize{$\pm$0.24}  &   60.43\scriptsize{$\pm$8.27}  &   31.41\scriptsize{$\pm$0.2}  &   23.94\scriptsize{$\pm$0.28}  &   13.63\scriptsize{$\pm$5.27}  &   3.16\scriptsize{$\pm$0.95}  &   2.1\scriptsize{$\pm$0.03}  &   \textbf{17.88\scriptsize{$\pm$5.09}}  &   6.07\scriptsize{$\pm$0.02} \\
    SimMatch & 5.55\scriptsize{$\pm$0.03}  &   5.51\scriptsize{$\pm$0.06}  &   4.64\scriptsize{$\pm$0.06}  &   42.17\scriptsize{$\pm$0.62}  &   30.2\scriptsize{$\pm$0.17}  &   23.77\scriptsize{$\pm$0.13}  &   14.4\scriptsize{$\pm$0.31}  &   3.89\scriptsize{$\pm$2.45}  &   2.15\scriptsize{$\pm$0.05}  &   27.95\scriptsize{$\pm$6.5}  &   6.39\scriptsize{$\pm$0.56} \\
    AdaMatch & 5.33\scriptsize{$\pm$0.22}  &   5.34\scriptsize{$\pm$0.05}  &   4.71\scriptsize{$\pm$0.02}  &   \textbf{40.19\scriptsize{$\pm$1.63}}  &   28.08\scriptsize{$\pm$0.39}  &   22.91\scriptsize{$\pm$0.18}  &   11.49\scriptsize{$\pm$3.52}  &   2.22\scriptsize{$\pm$0.06}  &   2.12\scriptsize{$\pm$0.08}  &   36.46\scriptsize{$\pm$5.49} &   6.43\scriptsize{$\pm$0.13} \\
    \midrule
    \rowcolor{LightGreen}\textbf{SequenceMatch}   &   \textbf{5.03\scriptsize{$\pm$0.11}}   &   \textbf{5.07\scriptsize{$\pm$0.11}}   &   \textbf{4.43\scriptsize{$\pm$0.02}}   &   44.52\scriptsize{$\pm$1.01} &   \textbf{27.16\scriptsize{$\pm$0.23}} &   22.90\scriptsize{$\pm$0.16}  &   \textbf{2.01\scriptsize{$\pm$0.43}}  &   \textbf{1.89\scriptsize{$\pm$0.01}}  &   \textbf{1.86\scriptsize{$\pm$0.01}}  &   40.21\scriptsize{$\pm$6.11}  &   \textbf{5.88\scriptsize{$\pm$0.14}} \\\bottomrule \bottomrule
    \end{tabular}%
\end{sc}
\end{small}
}
\end{table*}

\section*{Detailed results}
We also report the mean error rates of the last 20 checkpoints for various methods in Table \ref{table:mer}.
It can be seen that while most of the algorithms are overfitting to the training data at the end of the training process, our proposed method still maintains its robustness.

% \clearpage
\section*{ImageNet detailed results}
In this section, we show the detailed results of Table \ref{table:results2} for ImageNet dataset on 10\% labeled data.
We could see that SequenceMatch outperforms previous methods in both scenarios where self-supervised pre-trained weights are included or not.

\begin{table*}[!ht]
    \centering
    \caption{Accuracy results on ImageNet with 10\% labeled examples using \cite{li2021comatch} and \cite{zheng2022simmatch} source code.}
    \label{table:results3}
\begin{small}
\begin{sc}
    \resizebox{0.9\linewidth}{!}{
    \begin{tabular}{ll|cccc}
    \toprule \toprule
    Self-supervised &   \multirow{2}{*}{Method}         & \multirow{2}{*}{Top-1}          & \multirow{2}{*}{Top-5}          & Params &   \multirow{2}{*}{Epochs}             \\
    Pre-training    &   &           &           & (train/test)      &  \\
    \midrule \midrule
    None    &   FixMatch                                    & 71.5              & 89.1         & 25.6M/25.6M  &   $\sim300$              \\ 
    MoCo-EMAN\cite{cai2021exponential} &   FixMatch-EMAN\cite{cai2021exponential}     & 74.0              & 90.9         & 30.0M/25.6M  &   $\sim1100$               \\
    None &   CoMatch\cite{li2021comatch}                & 73.6              & 91.6         & 30.0M/25.6M  &   $\sim400$                \\ 
    MoCo V2\cite{chen2020improved} &   CoMatch\cite{li2021comatch}                & 73.7              & 91.4         & 30.0M/25.6M  &   $\sim1200$                \\ 
    None    &   SimMatch\cite{zheng2022simmatch}           & 74.4              & 91.6         & 30.0M/25.6M  &   $\sim400$             \\ 
    \midrule
    \rowcolor{LightGreen}None    &   \textbf{SequenceMatch}                                  & \textbf{75.2}     & \textbf{91.9}           & 25.6M/25.6M   &  $\sim400$     \\ \bottomrule \bottomrule
    \end{tabular}%
    }
\end{sc}
\end{small}
\end{table*}

\section*{Ablation study on medium augmentation}
\label{apx:medium}
For the medium one, we conduct ablation studies on various types of augmentation and report the results in Table \ref{table:augmentation}.
We systematically test different types of augmentation for medium one, such as MoCo \cite{he2020momentum}, SimCLR \cite{chen2020simple}, CTAugment \cite{Berthelot2020ReMixMatchSL}, and a combination of weak augmentation with 1 or 2 randomly selected strong augmentation.
As can be seen in Table \ref{table:augmentation}, using the combination of weak augmentation with 1 random strong augmentation results in the best performance.

% \begin{table*}[ht!]
% \centering
% \caption{Augmentation results on CIFAR-10 with 40-label split.}
% \begin{small}
% \begin{sc}
%     \begin{tabular}{l|c}
%     \toprule
%     Method                                                  &   Top-1\\
%     \midrule
%     \rowcolor{LightGreen}Weak augmentation + 1 random strong augmentation        &   \textbf{4.80}\\
%     Weak augmentation + 2 random strong augmentation        &   4.91\\
%     MoCo augmentation \cite{he2020momentum}                                       &   5.97\\
%     CTAugment \cite{Berthelot2020ReMixMatchSL}                                               &   4.85\\
%     SimCLR augmentation \cite{chen2020simple}                                          &   5.32\\    
%     \bottomrule
%     \end{tabular}%
% \end{sc}
% \end{small}
%     \label{table:augmentation}
% \end{table*}

\section*{List of data transformations}
\label{apx:augmentation}
We used the same sets of image transformations used in FixMatch \cite{sohn2020fixmatch}. 
For completeness, we listed all transformation operations for these augmentation strategies in Table \ref{table:randaugment}.

\begin{table*}[!ht]
    \centering
    \caption{List of transformations used in RandAugment}
    \label{table:randaugment}
\begin{small}
\begin{sc}
    \begin{tabular}{p{2.5cm}p{9cm}p{2cm}p{2cm}}
    \toprule \toprule
    Transformation & Description & Parameter & Range\\
    \midrule \midrule
    Autocontrast & Maximizes the image contrast by setting the darkest (lightest) pixel to black (white). & &\\
    Brightness & Adjusts the brightness of the image. $B = 0$ returns a black
    image, $B = 1$ returns the original image. & $B$ & [0.05, 0.95]\\
    Color & Adjusts the color balance of the image like in a TV. $C = 0$ returns a black \& white image, $C = 1$ returns the original image. & $C$ & [0.05, 0.95]\\
    Contrast & Controls the contrast of the image. A $C = 0$ returns a gray image, $C = 1$ returns the original image. & $C$ & [0.05, 0.95]\\
    Equalize & Equalizes the image histogram. & &\\
    Identity & Returns the original image. & &\\
    Posterize & Reduces each pixel to $B$ bits. & $B$ & [4, 8]\\
    Rotate & Rotates the image by $\theta$ degrees. & $\theta$ & [-30, 30]\\
    Sharpness & Adjusts the sharpness of the image, where $S = 0$ returns a blurred image, and $S = 1$ returns the original image. & $S$ & [0.05, 0.95]\\
    Shear\_x & Shears the image along the horizontal axis with rate $R$. & $R$ & [-0.3, 0.3] \\
    Shear\_y & Shears the image along the vertical axis with rate $R$. & $R$ & [-0.3, 0.3]\\
    Solarize & Inverts all pixels above a threshold value of $T$. & $T$ & [0, 1]\\
    Translate\_x & Translates the image horizontally by ($\lambda\times$image width) pixels. & $\lambda$ & [-0.3, 0.3]\\
    Translate\_y & Translates the image vertically by ($\lambda\times$image height) pixels. & $\lambda$ & [-0.3, 0.3] \\
    \bottomrule \bottomrule
    \end{tabular}
\end{sc}
\end{small}
\end{table*}

\begin{figure*}[!ht]
    \centering
    \subfloat[Weak]
    {\includegraphics[width=\linewidth]{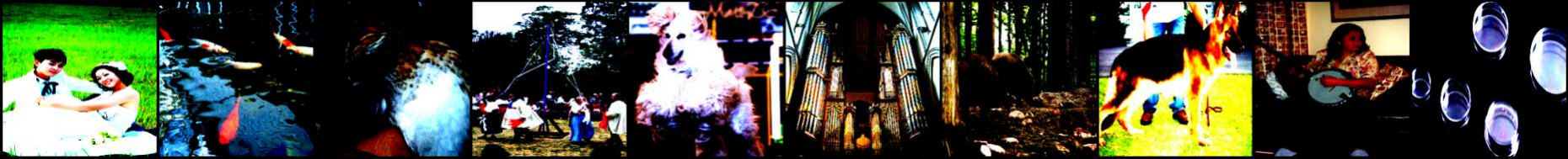}}% 
    \qquad
    \subfloat[Medium]
    {\includegraphics[width=\linewidth]{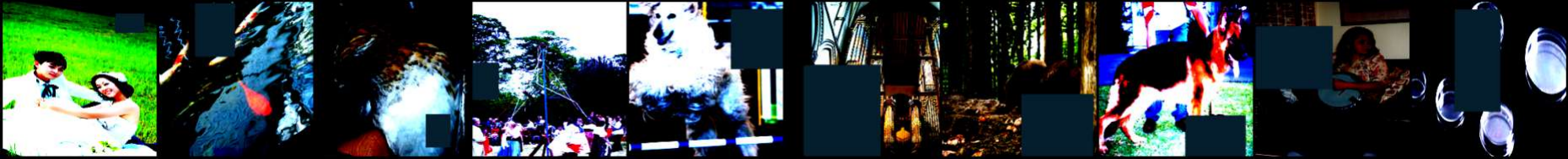}}%
    \qquad
    \subfloat[Strong]
    {\includegraphics[width=\linewidth]{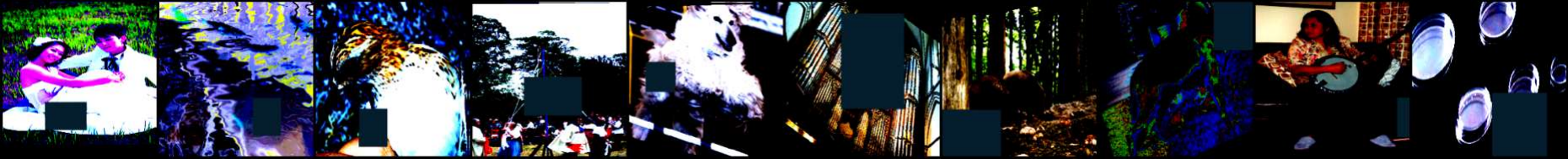}}%
    \caption{Weak, medium, and strong augmented examples.}
    \label{fig:augmented}
\end{figure*}

We visualize the weak, medium, and strong augmentation examples in Figure \ref{fig:augmented} for a better understanding of the differences among the three augmentations. 
As we can see, the mediumly augmented examples are different from the weakly augmented ones but they are not heavily distorted like the strongly augmented ones.

\section{Algorithm}
We present the complete algorithm for SequenceMatch in Algorithm \ref{alg:cap}.
\begin{algorithm*}[!ht]
\DontPrintSemicolon
  
    \KwInput{Labeled batch $\mathcal{X}={\left(x_{b},p_{b}\right):b\in{\left(1,\dots,B\right)}}$, unlabeled batch $\mathcal{U}={u_{b}:b\in{\left(1,\dots,\mu B\right)}}$, confidence threshold $\tau$, unlabeled data ratio $\mu$, unlabeled loss weight $\lambda_u$, temperature $\mathbf{T}$, $\Omega$ is $\mathcal{A}_w$, $\mathcal{A}_{m}$ or $\mathcal{A}_{s}$}
    \tcc{Cross-entropy loss for labeled data}
    $\mathcal{L}_{s}^{CE}=\frac{1}{B} \sum_{b=1}^{B} \mathrm{H}\left(p_{b}, \mathcal{A}_w\left(x_{b}\right)\right)$
    \For{$b = 1$ \KwTo $\mu B$}{
        $q_{b}\left(y \mid \Omega\left(u_{b}\right)\right)=p_{m}\left(y \mid \Omega\left(u_{b}\right)\right)$
        \tcp{Compute prediction after applying weak data augmentation of $u_{b}$}
        
        $q_{s} = \frac{\exp{\left(q_{b}/\mathbf{T}\right)}}{\sum_{k}\exp{\left(q_{k}/\mathbf{T}\right)}}$.     \tcp{Sharpen the output probability}  
    }
    \tcc{Cross-entropy loss with pseudo-label and confidence threshold for high-confidence unlabeled}
    $\mathcal{L}_{u}^\mathrm{CE} = \frac{1}{\mu B} \sum_{b=1}^{\mu B}\left(\mathbbm{1}\left(\max \left(q_{b}^{w}\right) \geq \tau\right) \mathrm{H}\left(\hat{q}_{b}, p_{m}\left(y \mid \mathcal{A}_{s}\left(u_{b}\right)\right)\right)+\mathbbm{1}\left(\max \left(q_{b}^{w}\right) < \tau\right)\mathrm{H}\left(q_s^{w}\mid q_{b}\left(\mathcal{A}_{s}\left(u_{b}\right)\right)\right)\right)$\\
    
    \tcc{Kullback-Leibler divergence loss with each pair of augmented examples}
    $\mathcal{L}_\mathrm{KL}^{w-m} = \frac{1}{\mu B} \sum_{b=1}^{\mu B}\mathbbm{1}\left(\max \left(q_{b}^{w}\right) \geq \tau\right)\qquad D_\mathrm{KL}\left(q_{s}^{w} \mid p_{m}\left(y \mid \mathcal{A}_{m}\left(u_{b}\right)\right)\right)$\\
    $\mathcal{L}_\mathrm{KL}^{m-s} = \frac{1}{\mu B} \sum_{b=1}^{\mu B}\mathbbm{1}\left(\max \left(q_{b}^{m}\right) \geq \tau\right)\qquad D_\mathrm{KL}\left(q_{s}^{m} \mid p_{m}\left(y \mid \mathcal{A}_{s}\left(u_{b}\right)\right)\right)$\\
    $\mathcal{L}_\mathrm{KL}^{w-s} = \frac{1}{\mu B} \sum_{b=1}^{\mu B}\mathbbm{1}\left(\max \left(q_{b}^{w}\right) \geq \tau\right)\qquad D_\mathrm{KL}\left(q_{s}^{w} \mid p_{m}\left(y \mid \mathcal{A}_{s}\left(u_{b}\right)\right)\right)$\\
    
    $\mathcal{L}_u = \mathcal{L}_{u}^\mathrm{CE} + \mathcal{L}_\mathrm{KL}^{w-m} + \mathcal{L}_\mathrm{KL}^{m-s} + \mathcal{L}_\mathrm{KL}^{w-s}$\\
    \Return{$\mathcal{L}_{s}^{CE}+\lambda_{u} \mathcal{L}_u$}
\caption{SequenceMatch algorithm}
\label{alg:cap}
\end{algorithm*}

\section{Long-tailed issue}
To further prove the effectiveness of SequenceMatch, we evaluate SequenceMatch on the imbalanced SSL setting.
We conduct experiments on CIFAR-10-LT, SVHN-LT, and CIFAR-100-LT with different imbalance ratios.
Following \cite{lee2021abc,wei2021crest,ren2020balanced}, we use WRN-28-2 as the backbone.
We consider long-tailed (LT) imbalance, where the number of data points exponentially decreases from the first class to the last, i.e., $N_k = N_1 \times \lambda^{-\frac{k-1}{L-1}}$, where $\lambda = \frac{N_1}{N_k}$.
For CIFAR-10, we set $\lambda=100, N_1=1000$, and $\beta=10\%, 20\%$, and $30\%$, respectively.
Similarly, we set $\lambda=100, N_1=1000$, and $\beta=20\%$ for SVHN.
And for CIFAR-100, we set $\lambda=20, N_1=200$, and $\beta=40$.
The results are recorded in Table \ref{table:results4-lt} with an average of three different runs.

\begin{table*}[ht!]
\centering
\caption{Overall accuracy under the long-tailed setting}
\label{table:results4-lt}
\resizebox{0.8\textwidth}{!}{
\begin{sc}
\begin{tabular}{lccccc}
\toprule    \toprule
                    & \multicolumn{3}{c}{CIFAR-10-LT}                               & SVHN-LT                   & CIFAR-100-LT \\ \midrule
\multirow{2}{*}{Algorithm}  & \multicolumn{3}{c}{$\lambda=100$}                     & $\lambda=100$             & $\lambda=20$ \\ 
                    & $\beta=10\%$      &   $\beta=20\%$  &  $\beta=30\%$           & $\beta=20\%$              & $\beta=40\%$ \\ \midrule  \midrule
Vanilla             & -                 & 55.3\scriptsize{$\pm$1.30}     & -                         &    77.0\scriptsize{$\pm$0.67}              &    40.1\scriptsize{$\pm$1.15} \\
VAT \cite{Miyato2019VirtualAT}           & -                 & 55.3\scriptsize{$\pm$0.88}     & -                         &    81.3\scriptsize{$\pm$0.47}              &    40.4\scriptsize{$\pm$0.34} \\
BALMS \cite{ren2020balanced}         & -                 & 70.7\scriptsize{$\pm$0.59}     & -                         &    87.6\scriptsize{$\pm$0.53}              &    50.2\scriptsize{$\pm$0.54} \\ \midrule
FixMatch \cite{sohn2020fixmatch}       & 70.0\scriptsize{$\pm$0.59}         & 72.3\scriptsize{$\pm$0.33}     & 74.9\scriptsize{$\pm$0.63}                 &    88.0\scriptsize{$\pm$0.30}              &    51.0\scriptsize{$\pm$0.20} \\
w/ CReST+PDA \cite{wei2021crest}  & 73.9\scriptsize{$\pm$0.40}         & 76.6\scriptsize{$\pm$0.46}     & 74.9\scriptsize{$\pm$0.63}                 &    89.1\scriptsize{$\pm$0.69}              &    51.6\scriptsize{$\pm$0.29} \\
w/ DARP \cite{kim2020distribution}       & -                 & 73.7\scriptsize{$\pm$0.98}     & -                         &    88.6\scriptsize{$\pm$0.19}              &    51.4\scriptsize{$\pm$0.37} \\
w/ DARP+cRT \cite{kim2020distribution}    & 74.6\scriptsize{$\pm$0.98}         & 78.1\scriptsize{$\pm$0.89}    & 77.6\scriptsize{$\pm$0.73}                &    89.9\scriptsize{$\pm$0.44}              &    54.7\scriptsize{$\pm$0.46} \\
w/ ABC \cite{lee2021abc}             & 77.2\scriptsize{$\pm$1.60}         & 81.1\scriptsize{$\pm$0.82}     & 81.5\scriptsize{$\pm$0.29}                 &    92.0\scriptsize{$\pm$0.38}              &    56.3\scriptsize{$\pm$0.19} \\
\rowcolor{LightGreen}\textbf{SequenceMatch} &  \textbf{85.4\scriptsize{$\pm$0.01}}                  & \textbf{81.5\scriptsize{$\pm$0.75}}     &      \textbf{82.2\scriptsize{$\pm$0.25}}                     &    \textbf{92.4\scriptsize{$\pm$0.06}}              &    \textbf{57.2\scriptsize{$\pm$0.09}} \\ \bottomrule   \bottomrule
\end{tabular}%
\end{sc}
}
\end{table*}

Surprisingly, SequenceMatch boosts the performance by a large margin when used with ABC \cite{lee2021abc}.
With an accuracy of 85.4\%, SequenceMatch outperforms ABC with an 8.2\% improvement when $\beta$ equals 10\%.

\end{document}